\documentclass[letterpaper]{article} 

\usepackage[]{aaai25}  
\usepackage{times}  
\usepackage{helvet}  
\usepackage{courier}  
\usepackage[hyphens]{url}  
\usepackage{graphicx} 
\usepackage{natbib}  
\usepackage{caption} 
\usepackage{algorithm}
\usepackage{algorithmic}
\usepackage[switch]{lineno}

\usepackage{newfloat}
\usepackage{listings}

\usepackage{amsmath}
\usepackage{amssymb}
\usepackage{amsfonts}

\usepackage[inline]{enumitem}
\usepackage{subfigure}
\usepackage{multirow}
\usepackage{float}
\usepackage{booktabs}
\usepackage{subcaption}
\usepackage{makecell}
\usepackage{xcolor}

\DeclareCaptionStyle{ruled}{labelfont=normalfont,labelsep=colon,strut=off} 
\floatstyle{ruled}
\newfloat{listing}{tb}{lst}{}
\floatname{listing}{Listing}
\title{Simulation-Free Hierarchical Latent Policy Planning for Proactive Dialogues}
\author{
    Tao He\textsuperscript{\rm 1}\thanks{Work was done during an internship at SMU.},
    Lizi Liao\textsuperscript{\rm 2},
    Yixin Cao\textsuperscript{\rm 3},
    Yuanxing Liu\textsuperscript{\rm 1},
    Yiheng Sun\textsuperscript{\rm 1},
    Zerui Chen\textsuperscript{\rm 1},
    Ming Liu\textsuperscript{\rm 1}\thanks{Corresponding Author: Ming Liu.},
    Bing Qin\textsuperscript{\rm 1}
}
\affiliations{
    \textsuperscript{\rm 1}Research Center for Social Computing and Information Retrieval, Harbin Institute of Technology, Harbin, China\\

    \textsuperscript{\rm 2}Singapore Management University, Singapore\\
    \textsuperscript{\rm 3}School of Computer Science, Fudan University\\
    \{the, yxliu, mliu, zrchen, qinb\}@ir.hit.edu.cn, lzliao@smu.edu.sg, caoyixin2011@gmail.com
}

\usepackage{bibentry}

\begin{document}

\maketitle

\begin{abstract}
Recent advancements in proactive dialogues have garnered significant attention, particularly for more complex objectives (e.g. emotion support and persuasion). Unlike traditional task-oriented dialogues, proactive dialogues demand advanced policy planning and adaptability, 
requiring rich scenarios and comprehensive policy repositories to develop such systems.
However, existing approaches tend to rely on Large Language Models (LLMs) for user simulation and online learning, leading to biases that diverge from realistic scenarios and result in suboptimal efficiency. Moreover, these methods depend on manually defined, context-independent, coarse-grained policies, which not only incur high expert costs but also raise concerns regarding their completeness.
In our work, we highlight the potential for automatically discovering policies directly from raw, real-world dialogue records.
To this end, we introduce a novel dialogue policy planning framework, LDPP. 
It fully automates the process from mining policies in dialogue records to learning policy planning. 
Specifically, we employ a variant of the Variational Autoencoder to discover fine-grained policies represented as latent vectors. After automatically annotating the data with these latent policy labels, we propose an Offline Hierarchical Reinforcement Learning (RL) algorithm in the latent space to develop effective policy planning capabilities. Our experiments demonstrate that LDPP outperforms existing methods on two proactive scenarios, even surpassing ChatGPT with only a 1.8-billion-parameter LLM. 
Our codes are available at \url{https://github.com/cs-holder/LDPP.git}.
\end{abstract}

\section{Introduction}
In recent years, there has been 
a surge of interest in dialogue tasks that require proactive engagement to achieve complex objectives, such as negotiation~\cite{He2018DecouplingSA}, persuasion~\cite{samad2022empathetic}, and emotional support~\cite{cheng2022improving}. Unlike traditional task-oriented dialogues~\cite{liu2022my, hu2023enhancing, liu2023mtgp}, these tasks require agents to be more proactive and possess sophisticated dialogue strategy skills~\cite{cheng2024cooper}. Previous research has demonstrated that even LLMs often struggle on such tasks~\cite{yang2021ubar, zhao2023chatgpt, kang2024can, song2024typing}. LLMs are typically trained to passively follow user instructions, which leads them to align with the user's opinions and decisions, lacking the necessary proactivity~\cite{deng2023plug, he2024planning}.

The advancement of LLMs in instruction-following and text generation capabilities has provided a foundation for exploring proactive dialogue systems, allowing a focus on high-level strategic research, i.e. dialogue policy planning~\cite{deng2023plug}, which plans the next dialogue policy to guide generating appropriate responses.
Some efforts have sought to directly enhance the 
strategic capabilities of LLMs by designing heuristic prompts or complex prompting processes~\cite{deng2023prompting, Yu2023PromptBasedMT}. 
However, these approaches often face limitations in performance or are criticized for high inference costs and inefficiency due to the need for continuous interactions. Other approaches aim to develop specialized policy planners to guide LLM responses strategically~\cite{deng2023plug}, allowing the separation of strategy from LLM and enabling a focused effort on learning policy planning capabilities. 

However, developing advanced policy planners requires rich exposure to \textbf{diverse dialogue scenarios} and access to a \textbf{comprehensive policy repository}.
Previous works~\cite{deng2023prompting} have used LLM like ChatGPT to simulate interactions, engaging in role-play and real-time learning. This methodology presents two critical drawbacks: first, the significant disparity between simulated and real-world interactions, as the toneless communication style of ChatGPT contrasts with the diverse and dynamic traits of actual human users; second, the reliance on continuous real-time interactions and frequent API calls for training, which introduces inefficiencies and escalates costs. Moreover, these approaches often depend on manually defined, context-independent, coarse-grained dialogue policies~\cite{zhou2019dynamic, liu2021towards}, which not only require substantial expert involvement but also raise concerns about the completeness and effectiveness of predefined policies. 

In this study, we introduce a novel paradigm that shifts away from relying on predefined policy sets and online learning in simulated environments, instead directly learning policy planning from raw, unlabeled dialogue records.
This paradigm effectively addresses two key challenges: 
1) It allows for \textbf{discovering fine-grained policies directly from realistic dialogues}, reducing the need for expert intervention and enhancing the completeness and relevance of resulting policies. 
2) By learning from real-world dialogues, it \textbf{eliminates the dependence on simulated environments}, thereby improving both efficiency and effectiveness.

To achieve this, we propose the innovative \textbf{Latent Dialogue Policy Planning (LDPP)} framework. LDPP automatically discovers policies as continuous latent vectors, expressing more semantics than predefined context-free policies, and facilitates the learning of effective planning within this latent policy space.
The framework consists of three key stages: \textbf{Latent policy discovery}, \textbf{Latent policy distillation}, and \textbf{Offline Hierarchical RL enhancement}. 
Inspired by the Variational Autoencoder’s (VAE) ability to encode inputs into an interpolable latent space~\cite{kingma2013auto}, we first employ a variant of the VQ-VAE~\cite{Oord2016PixelRN}
to automatically discover latent policies from dialogue records. These discovered latent policies are then used to label the training data. 
Finally, we propose an Offline Hierarchical Reinforcement Learning algorithm to both enhance the high-level policy planning and optimize response generation given latent policies at the lower token level.
Since the latent policies are represented as continuous vectors rather than natural language tokens, we further introduce the P-Former module. This module functions as a trainable adapter, ensuring that LLMs can effectively understand and follow the guidance of latent policies to respond, term as the latent-policy-following ability.
During inference, the policy planner first determines the appropriate latent policy based on the current dialogue state, which then directs the LLM in generating contextually relevant responses. 

To verify our approach, we conducted experiments widely on  ExTES~\cite{Zheng2023BuildingES}, ESConv~\cite{Liu2021TowardsES} and P4G~\cite{Wang2019PersuasionFG}. We compare our method with various baselines, demonstrating its effectiveness. Detailed analysis experiments further support the framework's validity. Our contributions are as follows:
\begin{itemize}
    \item We introduce a novel simulation-free dialogue policy planning learning framework, automatically mining potential policies from raw dialogue records.
    \item We propose an offline hierarchical reinforcement learning method for optimizing proactive dialogue, improving both planning capability and latent-policy-following ability for response generation.
    \item Extensive experiments across three proactive dialogue benchmarks show our approach outperforms baselines, with analysis confirming its effectiveness. 
\end{itemize}

\section{Related Work}
\noindent\textbf{Policy Planning for LLM-powered Dialogue Agent.} 
The advent of LLMs enables research into more complex dialogue tasks~\cite{cheng2024cooper} like emotion support and price negotiation. However, current studies indicate that LLMs often underperform in such tasks due to insufficient policy planning capacities~\cite{chen2023controllable}.
To improve policy planning, recent research has proposed various methods, which can be categorized into two parts: 
1) \textit{With predefined dialogue policy.} These methods need predefined dialogue policies, which can be further divided into two parts. Firstly, \citet{deng2023prompting} design a prompt process requiring LLMs to select an appropriate policy before generating a response.
GDP-Zero~\cite{Yu2023PromptBasedMT} employs Markov Monte Carlo Tree Search~\cite{Liebana2015OpenLS} to identify the next strategy. However, these methods are hindered by either the fixed parameters of LLMs or their high computational costs. To overcome this, PPDPP~\cite{deng2023plug} trains a specialized policy planner via online interaction with a simulated environment. \citet{zhang2024strength} increase richer user simulations to improve planning performance.
DPDP~\cite{he2024planning} employs the Dual-process theory~\cite{Kahneman2003MapsOB} to balance the efficiency and performance.
However, these methods require real-time interaction with a simulated environment, suffering from low efficiency and gaps between the realistic and simulated environment. 
2) \textit{Without predefined dialogue policy.} These approaches do not require pre-defined dialogue policies. Instead, they drive LLMs to analyze the current dialogue state and generate AI feedback, which is then used to help the LLMs to reply~\cite{fu2023improving, Zhang2023AskAE}. However, these methods often struggle to enhance the strategic reasoning capabilities of LLMs, resulting in less coherent and contextually appropriate responses, which leads to suboptimal performance. 

\noindent\textbf{Dialogue Generation on Latent Space.}
In the past years, studies have utilized latent features to control or enhance 
response generation~\cite{Wang2020ModellingHS, Cho2023DeepRW, Lubis2020LAVALA}. 
Some works employ VAE~\cite{Bowman2015GeneratingSF} variants such as CVAE~\cite{Zhao2017LearningDD}, 
and Discrete VAE~\cite{Bao2019PLATOPD} to model the semantic distribution of 
utterances in the latent space~\cite{Liu2020GoChatGC, Chen2022DialogVEDAP}, 
sampling latent variables to enhance response diversity~\cite{Xiang2024DiffusionDialogAD}. 
In this work, we focus on dialogue policy planning for LLM-powered proactive dialogues. We discover latent policies automatically and conduct planning within the latent space.

\section{Preliminaries}\label{sec:preliminaries}
\noindent\textbf{Problem formalization.}
Unlike previous works that focus solely on dialogue policy planning~\cite{deng2023plug}, our approach 
also optimizes the policy following ability for responding. 
To achieve this, we model the entire dialogue process using a hierarchical Markov Decision Process (MDP), inspired by recent studies 
\cite{zhou2024archer}. At the high level, a policy-level MDP is employed to model the policy planning task, while at the low level, a token-level MDP models the autoregressive generation of responses.

\begin{figure*}[t]
    \centering
    \includegraphics[width=\linewidth]{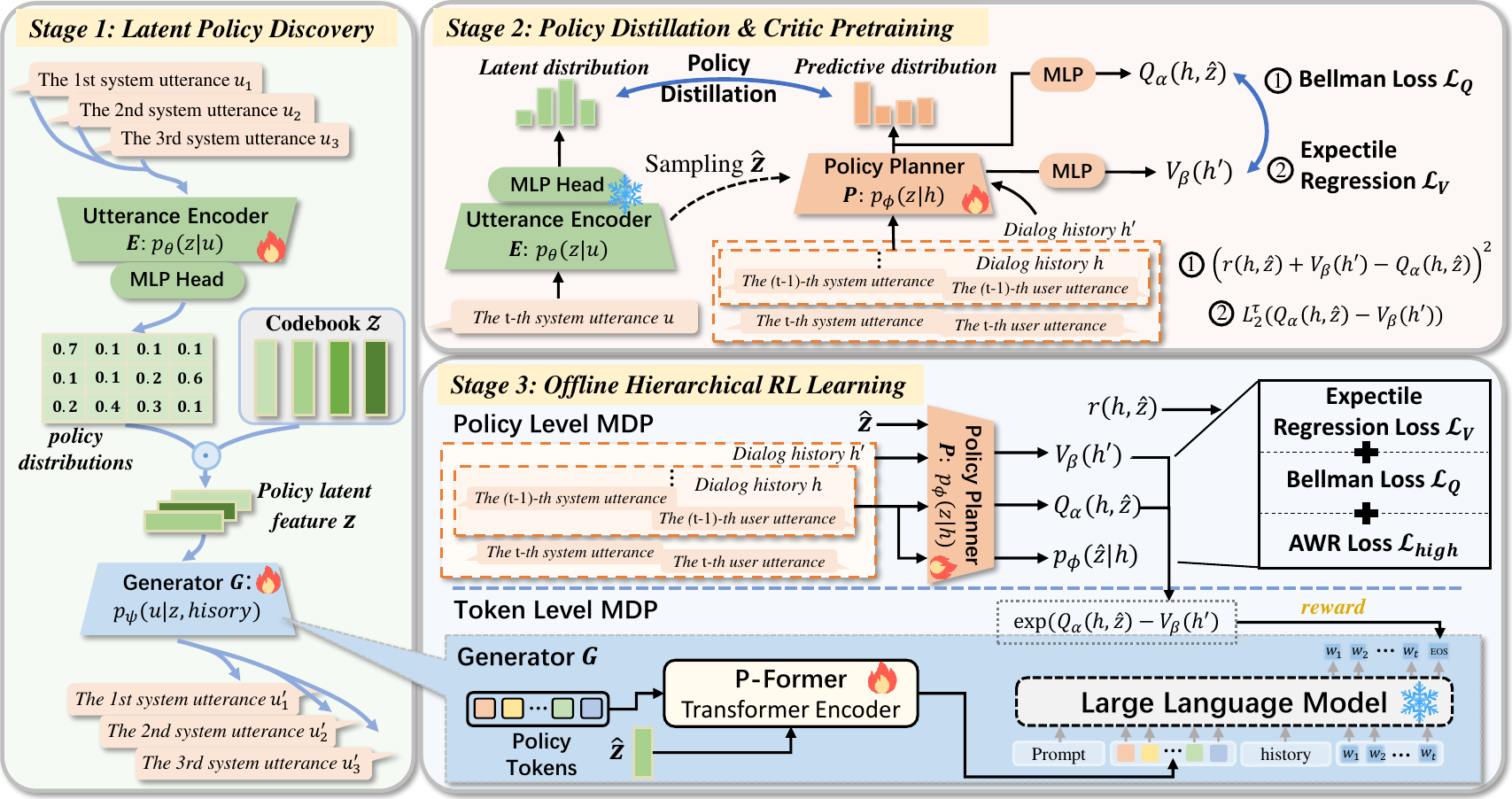}
    \caption{
    The training process of the LDPP framework.
    $u$ and $z$ refer to the system utterance and contained latent policy. $h$ and $h'$ denote $t$-th dialogue history $h_t$ and $(t+1)$-th dialogue history $h_{t+1}$, respectively.
    }
    \label{fig.main}
\end{figure*}

The policy-level MDP is defined as $\mathcal{M}_{h}=\left \langle \mathcal{S}_h, \mathcal{A}_h, \mathcal{R}_h\right \rangle$. The state set $\mathcal{S}_h$ consists of the dialogue history $h_t$ with alternating user utterances and system responses $\{u_1^{sys}, u_1^{usr}, \ldots, u_{t-1}^{sys}, u_{t-1}^{usr}\}$. 
The action $z_t\in\mathcal{A}_h$ refers to the dialogue policy, i.e., latent policy in this work. 
The reward function $\mathcal{R}_h$ evaluates each dialogue state using ChatGPT, outputting rewards $r_t$ for each turn of dialogue.
Please refer to the Evaluation Methods Section for details. 
Similarly, the token-level MDP is defined as $\mathcal{M}_l=\left \langle \mathcal{S}_l, \mathcal{A}_l, \mathcal{R}_l\right \rangle$, where the state set $\mathcal{S}_l=\{s_i^t=[h_t; z_t; w_{1:i-1}]\}$, with $w_{i}$ representing the $i$-th token of the response $u_t^{sys}=[w_1, w_2, \ldots, w_n]$. 
The action set $\mathcal{A}_l$ is the LLM's vocabulary, and the reward function $\mathcal{R}_l$ is provided by the policy-level MDP, detailed in the 3rd Stage introduction.
In the $t$-th dialogue turn, given the current state, i.e., dialogue history $h_t$, the policy planner predicts the appropriate dialogue policy $z_t$. 
Guided by the dialogue history $h_t$ and the dialogue policy $z_t$, the LLM generates the response $u_t^{sys}$.

In our proposed offline scenario, we only access raw dialogue records $\mathcal{D}$. For RL training, we decompose $\mathcal{D}$ into tuples: 
$\mathcal{D}=\{(h_t,u_t^{sys},u_t^{usr})\}$. 
To learn the policy-level MDP, we further use the policy-level reward $r_t$ for each dialogue turn $t$ to extend $\mathcal{D}$ as $\{(h_t,u_t^{sys},u_t^{usr},r_t)\}$.


\section{Component Models}

Before delving into the training framework, we first outline the component models. Our framework is composed of three key models: an \textbf{utterance encoder} $E$, a \textbf{policy planner} $P$, and a \textbf{generator} $G$. During the training phase, $E$ learns to discover latent policies from 
system responses and then annotate pseudo labels (latent policies) for training set for subsequent optimizing $P$ and $G$. 
In the inference phase, only $P$ and $G$ are actually employed: the planner $P$ outputs the next-turn policy based on the dialogue history, and then the policy is fed into $G$ to guide the response generation.

In this work, the design of the base models is not the central focus; therefore, we utilized RoBERTa-Large~\cite{Liu2019RoBERTaAR} as the base for both $E$ and $P$, same as works like PPDPP~\cite{deng2023plug}. $E$ takes system responses as input and uses the output of ``[CLS]'' to analysis the distribution of latent policies contained in responses; $P$ takes dialogue history as input and similarly outputs a predicted distribution of next-step policies. $G$ is based on an LLM.

However, LLMs only accept texts, while the latent policy is a continuous vector, which obviously has a significant gap between them. Inspired by the development of Vision Large Models~\cite{Li2023BLIP2BL}, we propose to train a \textbf{P-Former} to bridge this gap.
P-Former consists of $L$ stacked transformer layers, taking $T$ learnable policy tokens as input. These policy tokens interact with the latent policy features through a cross-attention mechanism. Ultimately, P-Former outputs $T$ policy-related tokens. We hope these tokens align with the input space of LLM, thus LLM can understand and follows the guidance of latent policies for appropriate response generation.
During training, the P-Former is optimized by the reconstruction loss of LLM. Notably, we freeze the LLM throughout. 
\textbf{Therefore, P-Former is also responsible for improving the latent-policy-following capacity of ${G}$.}

\section{Optimizaing Framework}
Our training framework is depicted in Figure~\ref{fig.main}. It consists of three stages with the following motivations and relationships: \textbf{Stage 1}: It focuses on automatically learning latent policies from raw dialogues. These latent policies serve as ``annotations'' for optimizing policy planning in Stage 3. \textbf{Stage 2}: It is used to initialize the policy planner, thereby accelerating and stabilizing the reinforcement learning process in Stage 3. \textbf{Stage 3}: Upon the preparatory work in Stage 1 and 2, this stage aims to enhance policy planning capabilities at the policy level and 
further optimize latent-policy-following abilities for responding at the token level.

\subsubsection{1st Stage: \textit{Latent Policy Discovery.}}\label{sec:1-stage}
We first automatically mine potential dialogue policies from raw dialogue records. The basic premise is that, given the dialogue history and the policy implied in one response, the dialogue agent should be able to reconstruct this response.
To this end, we propose an adjusted VQ-VAE algorithm. We first compress the inputted utterance into latent policy and then apply it, along with dialogue history, to guide the LLM in reconstructing the utterance. If the reconstruction is good, we assume the learned latent policy is effective. For more details about VQ-VAE, please refer to the appendix.

Like VQ-VAE, We define a codebook $\mathcal{Z} = \{\mathcal{Z}_k \in \mathbb{R}^d\}_{k=1}^K$ with $K$ policy vectors. Given a system utterance $u^{sys}_t$, shorted as $u_t$, we use the encoder $E$ to compress it and classify it into $K$ classes, yielding the policy distribution $p_\theta(z_t|u_t)\in\mathbb{R}^K$. Instead of performing a nearest neighbors lookup like VQ-VAE, we use $p_\theta(z_t|u_t)$ to perform a weighted sum of $\mathcal{Z}$ to obtain the latent policy feature $z_t$:
\begin{equation}
    z_t=\sum_{k=1}^K\mathcal{Z}_k\cdot p_{\theta,k}(z|u_t).
    \label{eq:policy_feature}
\end{equation}
This improvement allows us to involve multiple policies within a single response and expand the number of fine-grained policies through combinations.

For the generator $G$, the policy $z_t$ is first transferred into policy tokens using P-Former. Then policy tokens, along with the dialogue history $h_t$, are fed into the LLM to guide the generation of the response $u_t$.
By computing the reconstruction loss $\mathcal{L}_{con}$ of $G$ and propagating gradient backward, we can simultaneously optimize $E$, $G$, and $\mathcal{Z}$.


After the 1st training, we employ $E$ to annotate pseudo labels $\hat{z}_t$ for each system utterance in $\mathcal{D}$, expanding $\mathcal{D}$ to $\{(h_t, u_t^{sys}, u_t^{usr}, \hat{z}_t, r_t)\}$. Using $\mathcal{D}$, we are able to apply RL algorithm to optimize the policy planning capabilities.

\subsubsection{2nd Stage: \textit{Latent Policy Distillation.}}\label{sec:2-stage}
To expedite the RL training process in the 3rd stage, we initialize the policy planner $P$ by distilling the utterance encoder $E$.
Specifically, for a response $u_t$, we compute the predicted policy distributions using $E$ and $P$ as $p_\theta(z_t|u_t)$ and $p_\phi(z_t|h_t)$, respectively. Then we freeze $E$ and minimize the KL divergence to drive $P$ to learn from $E$.
However, we observe that the training set contains many inappropriate system utterances that lead to unsuccessful dialogues, which may harm $P$'s planning ability.
Therefore, we use the high-level rewards of each system response for data filtering, denoted as: 
\begin{equation}
    \begin{aligned}
        \mathcal{L}_{kl}(\phi)=\sum_{(u_t, h_t,r_t)\in\mathcal{D}}\mathbb{I}(r_t>\delta)\cdot\\
        \text{KL\_div}(p_\theta(z|u_t)||p_\phi(z|h_t)),
    \end{aligned}
\end{equation}
where $\mathbb{I}(.)$ refers to indicator function. $\theta$ and $\phi$ represent the trainable parameters of $E$ and $P$. And $\delta$ is a predefined threshold. We term this process policy distillation.

To stabilize RL learning, We also initialize the action-value function network $Q_\alpha$ and the value function network $V_\beta$ at this stage. These two networks evaluate dialogue states during Stage 3, which are implemented by stacking two MLP layers on the policy planner $P$. To pretrain $Q_\alpha$ and $V_\beta$, we use the off-the-shelf Offline RL algorithm IQL~\cite{Kostrikov2021OfflineRL}, with the following optimization objectives, respectively:
\begin{equation}
    \begin{aligned}
        \mathcal{L}_V(\beta)=\mathbb{E}_{(h_t,z_t)\in \mathcal{D}}[L_2^\tau(&Q_\alpha(h_t,z_t),V_\beta(h_t))],\\
        \mathcal{L}_Q(\alpha)=\mathbb{E}_{(h_t,z_t,H_{t+1})\in \mathcal{D}}&[(r_t + \gamma V_\beta(h_{t+1})\\
        &-Q_\alpha(h_t, z_t))^2],
        \label{eq:q_v_target}
    \end{aligned}
\end{equation}
where $\alpha$ and $\beta$ are trainable parameters of $Q_\alpha$ and $V_\beta$, respectively. And $\mathcal{L}_2^\tau$ means Expectile Regression Loss~\cite{Kostrikov2021OfflineRL}.
Therefore, the final optimization objective for this stage is as:
\begin{equation}
    \mathcal{L}_2=\mathcal{L}_{kl}(\phi)+\mathcal{L}_Q(\alpha)+\mathcal{L}_V(\beta).
\end{equation}

\subsubsection{3rd Stage: \textit{Offline Hierarchical RL Enhancement.}}\label{sec:3-stage}
To optimize this system only using training data without interactions with simulated environments, we propose an offline hierarchical RL algorithm to learn the policy-level and token-level MDPs simultaneously.

For the policy-level MDP, we utilize the IQL~\cite{Kostrikov2021OfflineRL} to simultaneously train the policy planner $P$, and the $Q$-, $V$-networks. The optimization objectives for the latter two are given by Eq.(\ref{eq:q_v_target}), and the optimization target for the policy planner $P$ is:
\begin{equation}
    \begin{aligned}
        \mathcal{L}_{high}(\phi)=-&\mathbb{E}_{(u_t,h_t,z_t)\sim\mathcal{D}}[\exp(\tau(Q_\alpha(h_t, z_t)\\
        &- V_\beta(h_t)))\log p_\phi(h_t|z_t)],
    \end{aligned}
\end{equation}
where $\tau\ge 0$ is the hyperparamter. The motivation behind the optimization target is to apply the advantage function $A(h_t, z_t) = Q_\alpha(h_t, z_t) - V_\beta(h_t)$ to weight each training sample $(u_t, h_t, z_t) \in \mathcal{D}$, thereby enabling selective learning from training data.

For the token-level MDP, we use the REINFORCE algorithm~\cite{Sutton1999PolicyGM} to optimize Generator $G$, aiming to improve the generation quality. Each intermediate token receives zero
reward, and a final reward of $\exp(A(h_t, z_t))$ is given after generating the complete $u_t$. We optimize $G$ using the following objective:
\begin{equation}
    \begin{aligned}
    \mathcal{L}_{low}(\psi)=-\sum_{(u_t,h_t,z_t)\sim\mathcal{D}}\exp{(A(h_t,z_t))}\\
    \cdot\sum_{w_i\in u_t}\log p_\psi(w_i|h_t,z_t,w_{1:i-1}),
    \end{aligned}
    \label{eq:L_low}
\end{equation}
where $\psi$ denotes the trainable parameters of Generator $G$.
For proof of this conclusion and empirical explanation, please refer to the appendix. 
It is important to note that we freeze the parameters of the LLM, so training Generator $G$ actually optimizes the P-Former. 
Ultimately, the training target of this stage is:
\begin{equation}
    \mathcal{L}_{3}=\mathcal{L}_{high}+ \mathcal{L}_{low}+\mathcal{L}_V+ \mathcal{L}_Q.
\end{equation}
By jointly training the policy planner $P$ and generator $G$, we simultaneously enhance the system's policy planning capability and the response quality given latent policies.

\section{Experimental Settings}
\subsection{Datasets}

\begin{table*}[t]
    \centering
    \renewcommand*{\arraystretch}{1.1}
    \resizebox{0.95\linewidth}{!}{
    \begin{tabular}{llcccccc|ccc}
    \toprule
    \multirow{2}{*}{\textbf{Policy Usage}} & \multirow{2}{*}{\textbf{Models}} & \multicolumn{3}{c}{\textbf{ExTES}} & \multicolumn{3}{c}{\textbf{Generalization to ESConv}} & \multicolumn{3}{c}{\textbf{P4G}}\\
    && SSR$\uparrow$ & SR$\uparrow$ & AvgT$\downarrow$ & SSR$\uparrow$ & SR$\uparrow$ & AvgT$\downarrow$ & SSR$\uparrow$ & SR$\uparrow$ & AvgT$\downarrow$ \\
    \midrule
    \multirow{3}{*}{Predefined Policy} & \multicolumn{1}{l}{Proactive} & 0.544 & 0.605 & 7.638 & 0.430 & 0.408 & 7.754 & 0.012 & 0.045 & 7.930 \\
    & \multicolumn{1}{l}{ProCoT} & 0.486 & 0.490 & 8.128 & 0.410 & 0.438 & 7.992 & 0.542 & 0.400 & 6.885 \\
    & \multicolumn{1}{l}{PPDPP} & 0.511 & 0.558 & 8.163 & 0.488 & 0.515 & 7.865 & 0.635 & \underline{0.745} & \underline{5.555} \\
    \midrule
    \multirow{6}{*}{No Need for Policy} & \multicolumn{1}{l}{Standard Prompt}  \\
    & \multicolumn{1}{l}{$\quad$+ ChatGPT} & \underline{0.650} & \underline{0.810} & \underline{6.138} & \underline{0.639} & \underline{0.762} & 6.546 & 0.477 & 0.460 & 7.025 \\
    & \multicolumn{1}{l}{$\quad$+ Qwen1.5-1.8b} & 0.538 & 0.613 & 7.590 & 0.543 & 0.623 & 6.723 & \underline{0.683} & 0.630 & {6.320} \\
    & \multicolumn{1}{l}{ICL-AIF} & 0.474 & 0.555 & 7.655 & 0.542 & 0.669 & \underline{6.415} & 0.063 & 0.070 & 7.640 \\
    & \multicolumn{1}{l}{LoRA Finetuning (32, 64)} & 0.558 & 0.627 & 7.308 & 0.616 & 0.662 & 6.738 & 0.651 & {0.655} & 6.645 \\
    & \multicolumn{1}{l}{LoRA Finetuning (64, 128)} & 0.566 & 0.628 & 7.450 & 0.583 & 0.654 & 6.892 & 0.541 & 0.570 & 6.830 \\
    \midrule
    \midrule
    \multirow{3}{*}{\makecell[l]{Automatically Discover\\Latent Policy}} & \multicolumn{1}{l}{LDPP} & \textbf{0.723} & \textbf{0.903} & \textbf{4.132} & \textbf{0.651} & \textbf{0.781} & \textbf{5.388} & \textbf{0.733} & \textbf{0.795} & \textbf{5.570} \\
    & \multicolumn{1}{l}{$\quad$-w/o \textit{2nd Stage}} & 0.716 & 0.865 & 4.483 & 0.637 & 0.769 & 5.608 & 0.715 & 0.760 & 6.140 \\
    & \multicolumn{1}{l}{$\quad$-w/o \textit{3rd Stage}} & 0.560 & 0.623 & 7.038 & 0.528 & 0.538 & 7.777 & 0.550 & 0.570 & 6.840 \\
    \bottomrule
    \end{tabular}
    }
    \caption{Main results on ExTES, ESConv, and P4G, using gpt-3.5-turbo-0613 as the critic. LoRA Fine-tuning(x, y) means setting \textit{lora rank}=x and \textit{lora alpha}=y. Results on ESConv are conducted using the planner trained on ExTES. }
    \label{tab:static_results}
\end{table*}


We evaluate the proposed framework on two typical applications of proactive dialogues, ExTES~\cite{zheng2023building} (emotional support) and P4G~\cite{wang2019persuasion} (persuasion), representing collaborative and non-collaborative dialogue, respectively. ExTES is an extension of ESConv~\cite{Liu2021TowardsES}, comprising sufficient dialogues for training (11,117 complete dialogues). We randomly divide it into 10,717/200/200 for train/valid/test set. P4G includes 1,017 donation persuasion dialogues where a ``persuader'' attempts to persuade a ``persuadee'' to donate to a charity called Save the Children. We randomly choose 100/100 dialogues for validation/testing. We take the remaining 817 dialogues as the training set. In practice, we extend the training set of dialogues to 5,579 using ChatGPT~\cite{Ouyang2022TrainingLM} due to the limited size. Please see the appendix for details of data augmentation. Given that ExTES is larger than P4G and P4G contains synthetic data, ExTES is more suitable for our task setup. Consequently, our primary analysis and experiments were conducted on ExTES. \textbf{Furthermore, to evaluate the generalizability of LDPP, we also test on ESConv (130 test cases) using LDPP trained on ExTES.}

\subsection{Baselines}
We compare Proactive~\cite{deng2023prompting}, ProCoT~\cite{deng2023prompting}, and PPDPP~\cite{deng2023plug} for baselines in need of predefined policies. 
Proactive and ProCoT require LLMs to select the most appropriate strategy before replying. PPDPP learns a specialized policy planner based on the predefined policies.
For methods not requiring policy use, we select the standard prompt method (prompting the base LLM to generate replies directly without considering dialogue policies), LoRA-based fine-tuning~\cite{Hu2021LoRALA} (shorted as LoRA), and ICL-AIF~\cite{fu2023improving}.
ICL-AIF prompts LLMs to provide suggestions before generating corresponding responses.

\subsection{Evaluation Methods}\label{lab:eval_methods}
\noindent\textbf{Self-play evaluation.} 
Since correct policies are often not unique and the absence of explicitly defined policies in our settings, directly assessing policy prediction accuracy is infeasible. 
We follow the same self-play method as previous work~\cite{deng2023plug} for dialogue-level evaluation. Specifically, two LLMs simulate the system and user in multi-turn dialogues, with the system receiving strategy guidance from a planner. We also prompt an LLM as critic to evaluate the completion status of each turn, deeming the dialogue failed if the goal isn't met within 10 turns. For more detailed prompts, please refer to the appendix.

\noindent\textbf{Critic model.} We also use ChatGPT to assess dialogue completion status following PPDPP. For ExTES and ESConv, we define four states: [worse, same, better, solved], with corresponding rewards of [-1, -0.5, 0.1, 1.0]; for P4G, the states are [reject, neutral, positive, donate], with also rewards of [-1, -0.5, 0.1, 1.0]. ChatGPT classifies the current dialogue state into one of 4 states. We perform 10 times classification per evaluation to reduce randomness, with each time getting a scalar value. We average them to obtain the reward $r_t$ for the current dialogue turn. 
A dialogue is considered successful if $r_t>\eta$. We set $\eta=0.6$ instead of 0.1 in PPDPP to improve the robustness of evaluations. 
\textbf{To ensure the robustness of results, we run the main experiments at least twice and reported the average results.}
\textbf{To further reduce evaluation bias, we use two versions of ChatGPT (gpt-3.5-turbo-0613 and -0125) to serve as critics and present the latter results in the appendix.}

\begin{table}[!t]
    \centering
    \subtable[ExTES]{
    \resizebox{0.9\linewidth}{!}{
    \begin{tabular}{lcccccccc}
    \toprule
    \textbf{LDPP} & \multicolumn{2}{c}{\textbf{Ide.}} & \multicolumn{2}{c}{\textbf{Com.}} & \multicolumn{2}{c}{\textbf{Sug.}} & \multicolumn{2}{c}{\textbf{Ove.}} \\
    \cmidrule(lr){2-3}\cmidrule(lr){4-5}\cmidrule(lr){6-7}\cmidrule(lr){8-9}
    vs. & Win & Lose & Win & Lose & Win & Lose & Win & Lose \\
    \midrule
    \textbf{PPDPP} & 8\% & 8\% & 52\% & 6\% & 64\% & 8\% & 68\% & 10\% \\
    \textbf{LoRA} & 6\% & 32\% & 6\% & 6\% & 32\% & 10\% & 26\% & 18\% \\
    \bottomrule
    \end{tabular}
    }
    }
    \subtable[P4G]{
    \resizebox{0.8\linewidth}{!}{
    \begin{tabular}{lcccccc}
    \toprule
    \textbf{LDPP} & \multicolumn{2}{c}{\textbf{Inf.}} & \multicolumn{2}{c}{\textbf{Per.}} & \multicolumn{2}{c}{\textbf{Ove.}} \\
    \cmidrule(lr){2-3}\cmidrule(lr){4-5}\cmidrule(lr){6-7}
    vs. & Win & Lose & Win & Lose & Win & Lose \\
    \midrule
    \textbf{PPDPP} & 32\% & 20\% & 40\% & 26\% & 48\% & 22\% \\
    \textbf{LoRA} & 10\% & 14\% & 24\% & 16\% & 26\% & 16\% \\
    \bottomrule
    \end{tabular}
    }
    }
    \caption{Human evaluation results on ExTES and P4G.}
    \label{tab:human_eval}
\end{table}

\noindent\textbf{Metrics.} Following \citet{deng2023plug}, we use two common dialogue-level metrics: Success Rate (SR) and Average Turn (AvgT). SR measures effectiveness and is defined as the ratio of the number of successful cases to the total number of test cases. AvgT measures the efficiency of goal completion by calculating the average dialogue turns of all test cases. 
However, we observe the high variance of SR. Therefore, we introduce the \textbf{SSR metric to more accurately assess effectiveness}.
SSR complements the SR, where SR calculates the ratio of success by mapping the final turn reward into a binary 0 or 1 while SSR averages all final turn rewards directly. Therefore, we view SSR as a ``Soft SR''.

\noindent\textbf{Backbone.} We conduct main experiments based on Qwen1.5-1.8b~\cite{qwen} and analysis studies on a series of LLMs: Qwen1.5-1.8b, -4b, -7b, Qwen2-1.5b, and Gemma-2b~\cite{Mesnard2024GemmaOM}. 
Due to the hardware limitations, we select models under 7B parameters.
We employ these LLMs to play the roles of Therapist/Persuader, respectively, guided by policies from the planner.

\section{Results and Analysis}
\subsection{Main Results}
Based on Table~\ref{tab:static_results}, we find that \emph{LDPP outperforms all baselines significantly on all tasks}.
This LDPP is implemented with $(T, L, K)=(8, 6, 24)$.
Firstly, LDPP achieves notable enhancements compared to the standard prompt and LoRA methods, verifying the effectiveness of latent policies and the P-Former module.
Prompt-based methods like Proactive, ProCoT, and ICL-AIF show unsatisfactory and unstable performance. We observe serious role confusion issues in these works. Due to the disturbance of suggestions or analyses, the system's responses fail to meet the expected form, leading to the role confusion during dialogue. We attribute this to the limited instruction-following and analysis capabilities of the 1.8b LLM.
Compared to PPDPP, LDPP performs more effectively and more efficiently without online learning and predefined policies, proving the effectiveness of self-supervised policy discovery and offline hierarchical RL training method.
Besides, we also find that \emph{LDPP based on Qwen1.5-1.8B performs better than ChatGPT}, further affirming our method's effectiveness. 
This also demonstrates that, with the assistance of external modules, smaller LLMs can surpass larger ones. 
For more results with the different LLM as critic, please refer to Table~\ref{tab:main_results_2} in the appendix.

Furthermore, we conduct ablation experiments by skipping Stage 2 and Stage 3. 
Firstly, the significant performance drop without Stage 3 underscores its necessity for learning policy planning. Without Stage 3, the policy planner can only learn from the utterance encoder, thus failing to acquire planning capabilities. Besides, the slight decline observed without Stage 2 also shows the rationality for proper initialization for effective RL-based policy planning.

\subsection{Human Evaluation}
Following previous studies~\cite{he2024planning}, we conduct human evaluation on 50 dialogues randomly sampled from the test in ExTES and P4G, respectively. We selected two training-based baselines, PPDPP and LoRA, based on whether they require predefined policies and a simulated environment.
Three annotators are required to compare the dialogues generated by LDPP/PPDPP and LDPP/LoRA. We assess four metrics: \textbf{Identification (Ide.)}, \textbf{Comforting (Com.)}, \textbf{Suggestion (Sug.)}, and \textbf{Overall (Ove.)} for ExTES and three metrics: \textbf{Information (Inf,)}, \textbf{Persuasion (Per.)}, and \textbf{Overall (Ove.)} for P4G. Detailed instructions for the annotators are provided in the appendix. Results are presented in Table~\ref{tab:human_eval}. 
First, LDPP outperforms PPDPP and LoRA in the \textbf{Ove.} metric, aligning with results in Table~\ref{tab:static_results}. 
We observe that LDPP does not like to ask patients for specific details, often providing suggestions quickly after the patient's introduction. 
While providing useful suggestions is crucial and could improve SR evaluation, failing to conduct thorough inquiries impacts the practical experience.
To alleviate this phenomenon, designing relevant rewards could be helpful. 

\subsection{Performance on Different LLMs}
\begin{figure}[t]
    \centering
    \subfigure[LLM Size]{
        \includegraphics[width=0.473\linewidth]{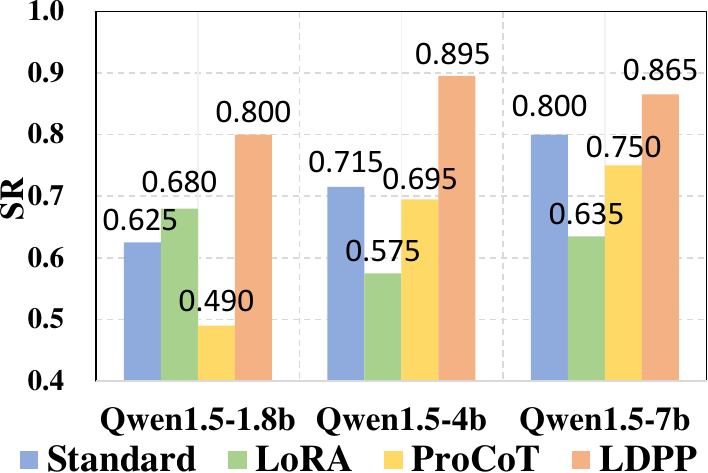}
    }
    \subfigure[LLM Series]{
        \includegraphics[width=0.473\linewidth]{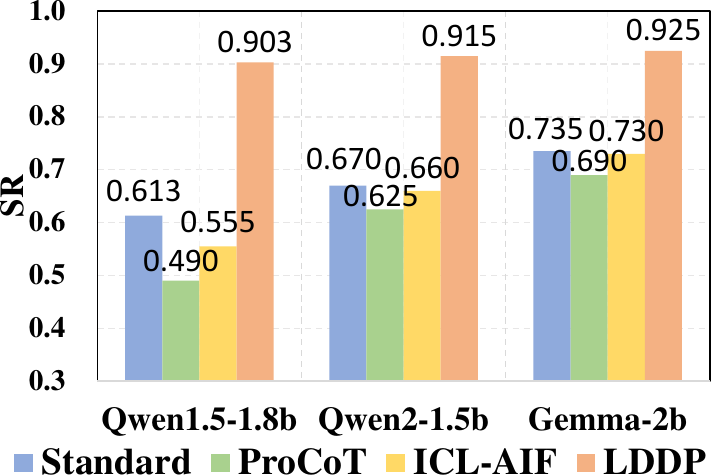}
    }
    \caption{Performance comparison as the LLM size and LLM series change on ExTES.} 
    \label{fig:llm_size}
\end{figure}

To further validate our proposed framework, we conduct experiments on LLMs with different sizes. Specifically, we compare LDPPs based on Qwen1.5-1.8b, 4b, and 7b for different sizes with settings of $(T,L,K)$=$(8,4,24)$. 
The results are shown in Figure~\ref{fig:llm_size}.
We observe that \emph{LDPP achieves the best performance in all three different sizes}.
As LLM size increases, standard prompting and prompting-based method ProCoT show continuous improvement, but they still perform worse than LDPP.
In contrast, LoRA Fine-tuning exhibits significant variability.
The reason may be that fine-tuning fails to differentiate data quality and train the added parameters sufficiently, harming LLMs' generalization ability.
Besides, we also conduct experiments ($(T, L, K)$=$(8, 6, 24)$) using Qwen1.5-1.8b, Qwen2-1.5b, and Gemma-2b for different LLM series and present in Figure~\ref{fig:llm_size}, we find that LDPP also performs best.

\subsection{Latent Policy Visualization}
\begin{figure}[!t]
    \centering
    \subfigure[ExTES]{
        \includegraphics[width=0.9\linewidth]{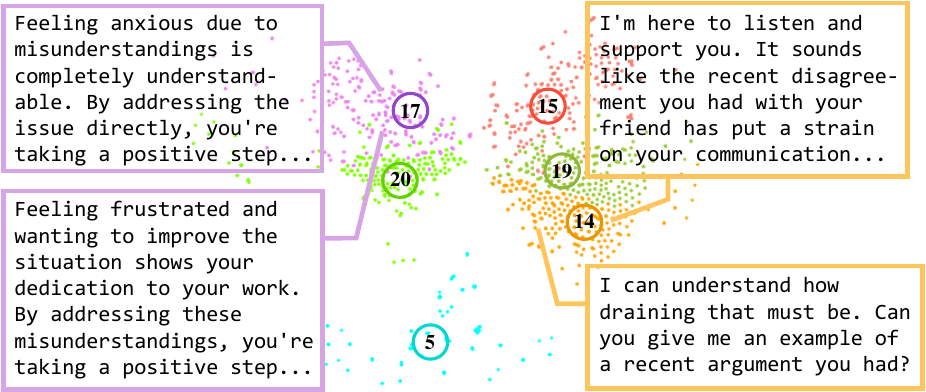}
    }
    \subfigure[P4G]{
        \includegraphics[width=0.95\linewidth]{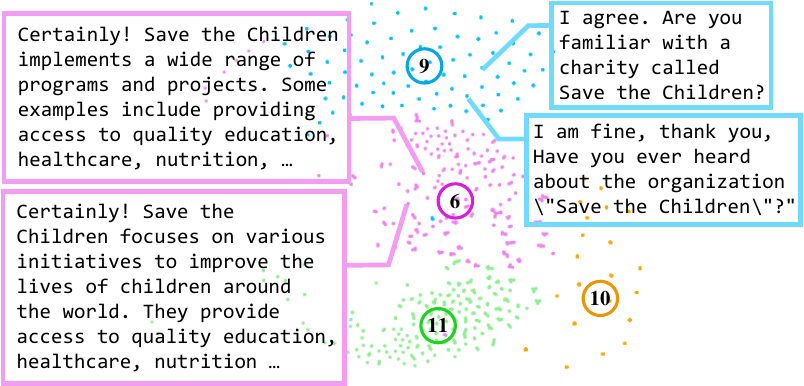}
    }
    \caption{Visualization of latent policies for utterances belong to top-4 and top-6 most frequently used policies.}
    \label{fig:visual}
\end{figure}

To intuitively demonstrate the learned latent policies, we visualize the policies of system utterances in Figure~\ref{fig:visual}.
Initially, each utterance is encoded into a latent policy feature following Eq.(\ref{eq:policy_feature}) and classified into the closest policy vector in the Codebook.
For each policy vector in the Codebook, we select the top-500 closest latent policy features and then apply PCA for dimensionality reduction on them to draw a scatter plot. For ease of presentation, we only display those from the 6/4 most frequently used policy vectors for ExTES/P4G. 
We also present parts of text utterances for comparison and observe that utterances within the same cluster are indeed semantically similar, validating the effectiveness of stage 1. To better understand these policies, Table~\ref{tab:example_utterances_extes} in the appendix presents three representative utterances for each of them. These utterance examples can help to understand the semantical operations for policies in the Codebook.

\begin{table}[t]
    \centering
    \resizebox{0.7\linewidth}{!}{
    \begin{tabular}{lcccc}
    \toprule
    & \textit{K}=6 & \textit{K}=12 & \textit{K}=18 & \textit{K}=24 \\
    \midrule
    \textbf{SSR} & 0.687 & 0.652 & 0.675 & 0.628 \\
    \textbf{AvgT} & 4.50 & 5.84 & 4.86 & 5.77 \\
    \bottomrule
    \end{tabular}
    }
    \caption{Results of different $K$ on ExTES.}
    \vspace{-0.2cm}
    \label{tab:num_policy}
\end{table}

\begin{table}[t]
    \centering
    \resizebox{0.7\linewidth}{!}{
    \begin{tabular}{lcccc}
    \toprule
    & \textit{T}=2 & \textit{T}=8 & \textit{T}=16 & \textit{T}=24 \\
    \midrule
    \textbf{SSR} & 0.699 & 0.628 & 0.619 & 0.628 \\
    \textbf{AvgT} & 4.57 & 5.77 & 6.17 & 5.65 \\
    \bottomrule
    \end{tabular}
    }
    \caption{Results of different \#policy tokens ($T$) on ExTES.}
    \vspace{-0.2cm}
    \label{tab:query_token}
\end{table}


\begin{table}[t]
    \centering
    \resizebox{0.9\linewidth}{!}{
    \begin{tabular}{lcccccc}
    \toprule
    & \multicolumn{3}{c}{\textbf{ExTES}} & \multicolumn{2}{c}{\textbf{P4G}} \\
    \cmidrule(lr){2-4} \cmidrule(lr){5-7}
    & \textit{L}=2 & \textit{L}=4 & \textit{L}=6 & \textit{L}=2 & \textit{L}=4 & \textit{L}=6 \\
    \midrule
    \textbf{SSR} & 0.649 & 0.628 & 0.719 & 0.580 & 0.711 & 0.732 \\
    \textbf{AvgT} & 5.42 & 5.77 & 3.88 & 6.37 & 5.85 & 5.49 \\
    \bottomrule
    \end{tabular}
    }
    \caption{Results of different P-Former layers ($L$).}
    \vspace{-0.2cm}
    \label{tab:policy_layer}
\end{table}

\subsection{Parameter Sensitivity Analysis}
\noindent\textbf{Codebook Size $K$.}
We investigate the impact of Codebook size $K$ on guiding the proactive dialogue process. Experiments are conducted on the ExTES dataset with $K$= 6, 12, 18, and 24, while keeping other hyper-parameters constant $(T=8, L=4)$, as shown in Table~\ref{tab:num_policy}. 
LDPP achieves relatively stable results and performs satisfactorily even with the smallest $K$, which can be attributed to the method of capturing latent policy features: by computing a weighted sum of the Codebook based on the policy distribution derived from the policy planner, it allows for a semantic combination of different policy vectors within the Codebook. Therefore, even with a small $K$, a wide range of latent policies can be expressed. However, performance decreased when $K=24$. We assume that this is due to the increased complexity of predicting the distribution for the larger Codebook, which requires additional training steps.

\noindent\textbf{\#Policy Tokens $T$.} Although we aim for these policy tokens to align with the input word embeddings of LLMs, they do not inherently belong to the LLMs' vocabulary. Therefore, it is important to analyze the potential noise introduced by the policy tokens into the LLMs. We set $T$ as 2, 8, 16, and 24 while keeping $(L=4, K=24)$. The experimental results are presented in Table~\ref{tab:query_token}. Overall, there is a trend of decreasing dialogue success rate as $T$ increases, indicating that a greater number of policy tokens indeed introduce noise, adversely affecting response generation. In practice, users can reduce the number of query tokens or enhance the capacity of P-Former (e.g., increasing the number of layers) to mitigate the impact of noise.

\noindent\textbf{P-Former Layer $L$.}
The number of P-Former layers reflects its parameter size and capability. We hypothesize that a stronger P-Former reduces the gap between the transferred policy tokens and the LLMs' input space while retaining more policy semantic information. To validate this, we set different layers on ExTES and P4G. The results, presented in Table~\ref{tab:policy_layer}, indicate that the number of P-Former layers impacts dialogue performances, especially on P4G, where more layers notably improve the dialogue success rate. On the P4G dataset, we observed zero improvement.
This indicates that only if the P-Former is sufficiently powerful can we effectively utilize the latent policy.

\section{Conclusion and Future Work}
In this work, we introduce a novel learning scenario that discovers potential policies from broadly collected dialogue records and learns policy planning without dynamic interactions with simulated environments. To address this challenge, we propose a new learning framework called LDPP, containing three stages:
latent policy discovery, policy distillation, and offline RL enhancement.
Experimental results demonstrate that LDPP significantly improves LLMs' proactive dialogue capabilities, achieving more pronounced and consistent enhancements compared to all baselines, even ChatGPT. 
Future research will mainly focus on improving the explainability of latent policies, ensuring the reliability of policies used in proactive dialogue.

\section*{Acknowledgements}
The research in this article is supported by the National Science Foundation of China (U22B2059, 62276083), the Human-Machine Integrated Consultation System for Cardiovascular Diseases (2023A003). We also appreciate the support from China Mobile Group Heilongjiang Co., Ltd. @ on our research,the research is jointly completed by both parties. This work was also supported by the Singapore Ministry of Education (MOE) Academic Research Fund (AcRF) Tier 1 grant (Proposal ID: 23-SIS-SMU-010). We are sincerely grateful to all reviewers for their insightful feedback.


\bibliography{aaai25}

\begin{thebibliography}{48}
\providecommand{\natexlab}[1]{#1}

\bibitem[{Bai et~al.(2023)Bai, Bai, Chu, Cui, Dang, Deng, Fan, Ge, Han, Huang, Hui, Ji, Li, Lin, Lin, Liu, Liu, Lu, Lu, Ma, Men, Ren, Ren, Tan, Tan, Tu, Wang, Wang, Wang, Wu, Xu, Xu, Yang, Yang, Yang, Yang, Yao, Yu, Yuan, Yuan, Zhang, Zhang, Zhang, Zhang, Zhou, Zhou, Zhou, and Zhu}]{qwen}
Bai, J.; Bai, S.; Chu, Y.; Cui, Z.; Dang, K.; Deng, X.; Fan, Y.; Ge, W.; Han, Y.; Huang, F.; Hui, B.; Ji, L.; Li, M.; Lin, J.; Lin, R.; Liu, D.; Liu, G.; Lu, C.; Lu, K.; Ma, J.; Men, R.; Ren, X.; Ren, X.; Tan, C.; Tan, S.; Tu, J.; Wang, P.; Wang, S.; Wang, W.; Wu, S.; Xu, B.; Xu, J.; Yang, A.; Yang, H.; Yang, J.; Yang, S.; Yao, Y.; Yu, B.; Yuan, H.; Yuan, Z.; Zhang, J.; Zhang, X.; Zhang, Y.; Zhang, Z.; Zhou, C.; Zhou, J.; Zhou, X.; and Zhu, T. 2023.
\newblock Qwen Technical Report.
\newblock \emph{arXiv preprint arXiv:2309.16609}.

\bibitem[{Bao et~al.(2019)Bao, He, Wang, and Wu}]{Bao2019PLATOPD}
Bao, S.; He, H.; Wang, F.; and Wu, H. 2019.
\newblock PLATO: Pre-trained Dialogue Generation Model with Discrete Latent Variable.
\newblock In \emph{Annual Meeting of the Association for Computational Linguistics}.

\bibitem[{Bowman et~al.(2015)Bowman, Vilnis, Vinyals, Dai, J{\'o}zefowicz, and Bengio}]{Bowman2015GeneratingSF}
Bowman, S.~R.; Vilnis, L.; Vinyals, O.; Dai, A.~M.; J{\'o}zefowicz, R.; and Bengio, S. 2015.
\newblock Generating Sentences from a Continuous Space.
\newblock In \emph{Conference on Computational Natural Language Learning}.

\bibitem[{Chen et~al.(2023)Chen, Yu, Shi, Awasthi, and Yu}]{chen2023controllable}
Chen, M.; Yu, X.; Shi, W.; Awasthi, U.; and Yu, Z. 2023.
\newblock Controllable mixed-initiative dialogue generation through prompting.
\newblock \emph{arXiv preprint arXiv:2305.04147}.

\bibitem[{Chen et~al.(2022)Chen, Gong, Wang, Yao, Qi, Wei, Hu, Zhou, Mao, Chen, Cheng, and Duan}]{Chen2022DialogVEDAP}
Chen, W.; Gong, Y.; Wang, S.; Yao, B.; Qi, W.; Wei, Z.; Hu, X.-M.; Zhou, B.; Mao, Y.; Chen, W.; Cheng, B.; and Duan, N. 2022.
\newblock DialogVED: A Pre-trained Latent Variable Encoder-Decoder Model for Dialog Response Generation.
\newblock In \emph{Annual Meeting of the Association for Computational Linguistics}.

\bibitem[{Cheng et~al.(2022)Cheng, Liu, Li, Wang, Zhao, Liu, Liang, and Zheng}]{cheng2022improving}
Cheng, Y.; Liu, W.; Li, W.; Wang, J.; Zhao, R.; Liu, B.; Liang, X.; and Zheng, Y. 2022.
\newblock Improving multi-turn emotional support dialogue generation with lookahead strategy planning.
\newblock \emph{arXiv preprint arXiv:2210.04242}.

\bibitem[{Cheng et~al.(2024)Cheng, Liu, Wang, Leong, Ouyang, Li, Wu, and Zheng}]{cheng2024cooper}
Cheng, Y.; Liu, W.; Wang, J.; Leong, C.~T.; Ouyang, Y.; Li, W.; Wu, X.; and Zheng, Y. 2024.
\newblock Cooper: Coordinating specialized agents towards a complex dialogue goal.
\newblock In \emph{Proceedings of the AAAI Conference on Artificial Intelligence}, volume~38, 17853--17861.

\bibitem[{Cho et~al.(2023)Cho, Takahashi, Yanase, and Saito}]{Cho2023DeepRW}
Cho, I.; Takahashi, R.; Yanase, Y.; and Saito, H. 2023.
\newblock Deep RL with Hierarchical Action Exploration for Dialogue Generation.
\newblock \emph{ArXiv}, abs/2303.13465.

\bibitem[{Deng et~al.(2023{\natexlab{a}})Deng, Lei, Liao, and Chua}]{deng2023prompting}
Deng, Y.; Lei, W.; Liao, L.; and Chua, T.-S. 2023{\natexlab{a}}.
\newblock Prompting and Evaluating Large Language Models for Proactive Dialogues: Clarification, Target-guided, and Non-collaboration.
\newblock \emph{arXiv preprint arXiv:2305.13626}.

\bibitem[{Deng et~al.(2023{\natexlab{b}})Deng, Zhang, Lam, Ng, and Chua}]{deng2023plug}
Deng, Y.; Zhang, W.; Lam, W.; Ng, S.-K.; and Chua, T.-S. 2023{\natexlab{b}}.
\newblock Plug-and-play policy planner for large language model powered dialogue agents.
\newblock In \emph{The Twelfth International Conference on Learning Representations}.

\bibitem[{Fu et~al.(2023)Fu, Peng, Khot, and Lapata}]{fu2023improving}
Fu, Y.; Peng, H.; Khot, T.; and Lapata, M. 2023.
\newblock Improving language model negotiation with self-play and in-context learning from ai feedback.
\newblock \emph{arXiv preprint arXiv:2305.10142}.

\bibitem[{He et~al.(2018)He, Chen, Balakrishnan, and Liang}]{He2018DecouplingSA}
He, H.; Chen, D.; Balakrishnan, A.; and Liang, P. 2018.
\newblock Decoupling Strategy and Generation in Negotiation Dialogues.
\newblock \emph{ArXiv}, abs/1808.09637.

\bibitem[{He et~al.(2024)He, Liao, Cao, Liu, Liu, Chen, and Qin}]{he2024planning}
He, T.; Liao, L.; Cao, Y.; Liu, Y.; Liu, M.; Chen, Z.; and Qin, B. 2024.
\newblock Planning Like Human: A Dual-process Framework for Dialogue Planning.
\newblock \emph{arXiv preprint arXiv:2406.05374}.

\bibitem[{Hu et~al.(2021)Hu, Shen, Wallis, Allen-Zhu, Li, Wang, and Chen}]{Hu2021LoRALA}
Hu, J.~E.; Shen, Y.; Wallis, P.; Allen-Zhu, Z.; Li, Y.; Wang, S.; and Chen, W. 2021.
\newblock LoRA: Low-Rank Adaptation of Large Language Models.
\newblock \emph{ArXiv}, abs/2106.09685.

\bibitem[{Hu et~al.(2023)Hu, Feng, Deng, Li, Ng, Luu, and Hooi}]{hu2023enhancing}
Hu, Z.; Feng, Y.; Deng, Y.; Li, Z.; Ng, S.-K.; Luu, A.~T.; and Hooi, B. 2023.
\newblock Enhancing Large Language Model Induced Task-Oriented Dialogue Systems Through Look-Forward Motivated Goals.
\newblock \emph{arXiv preprint arXiv:2309.08949}.

\bibitem[{Kahneman(2003)}]{Kahneman2003MapsOB}
Kahneman, D. 2003.
\newblock Maps of Bounded Rationality: Psychology for Behavioral Economics.
\newblock \emph{The American Economic Review}, 93: 1449--1475.

\bibitem[{Kang et~al.(2024)Kang, Kim, Kwon, Moon, Cho, Yu, Lee, and Yeo}]{kang2024can}
Kang, D.; Kim, S.; Kwon, T.; Moon, S.; Cho, H.; Yu, Y.; Lee, D.; and Yeo, J. 2024.
\newblock Can Large Language Models be Good Emotional Supporter? Mitigating Preference Bias on Emotional Support Conversation.
\newblock \emph{arXiv preprint arXiv:2402.13211}.

\bibitem[{Kingma and Welling(2013)}]{kingma2013auto}
Kingma, D.~P.; and Welling, M. 2013.
\newblock Auto-encoding variational bayes.
\newblock \emph{arXiv preprint arXiv:1312.6114}.

\bibitem[{Kostrikov, Nair, and Levine(2021)}]{Kostrikov2021OfflineRL}
Kostrikov, I.; Nair, A.; and Levine, S. 2021.
\newblock Offline Reinforcement Learning with Implicit Q-Learning.
\newblock \emph{ArXiv}, abs/2110.06169.

\bibitem[{Li et~al.(2023)Li, Li, Savarese, and Hoi}]{Li2023BLIP2BL}
Li, J.; Li, D.; Savarese, S.; and Hoi, S. C.~H. 2023.
\newblock BLIP-2: Bootstrapping Language-Image Pre-training with Frozen Image Encoders and Large Language Models.
\newblock In \emph{International Conference on Machine Learning}.

\bibitem[{Liebana et~al.(2015)Liebana, Dieskau, Hunermund, Mostaghim, and Lucas}]{Liebana2015OpenLS}
Liebana, D.~P.; Dieskau, J.; Hunermund, M.; Mostaghim, S.; and Lucas, S. M.~M. 2015.
\newblock Open Loop Search for General Video Game Playing.
\newblock \emph{Proceedings of the 2015 Annual Conference on Genetic and Evolutionary Computation}.

\bibitem[{Liu et~al.(2023)Liu, Wang, Tan, Zhao, Huang, He, and Hou}]{liu2023mtgp}
Liu, A.; Wang, B.; Tan, Y.; Zhao, D.; Huang, K.; He, R.; and Hou, Y. 2023.
\newblock MTGP: Multi-turn Target-oriented Dialogue Guided by Generative Global Path with Flexible Turns.
\newblock In \emph{Findings of the Association for Computational Linguistics: ACL 2023}, 259--271.

\bibitem[{Liu, Pan, and Luo(2020)}]{Liu2020GoChatGC}
Liu, J.; Pan, F.; and Luo, L. 2020.
\newblock GoChat: Goal-oriented Chatbots with Hierarchical Reinforcement Learning.
\newblock \emph{Proceedings of the 43rd International ACM SIGIR Conference on Research and Development in Information Retrieval}.

\bibitem[{Liu et~al.(2021{\natexlab{a}})Liu, Zheng, Demasi, Sabour, Li, Yu, Jiang, and Huang}]{liu2021towards}
Liu, S.; Zheng, C.; Demasi, O.; Sabour, S.; Li, Y.; Yu, Z.; Jiang, Y.; and Huang, M. 2021{\natexlab{a}}.
\newblock Towards emotional support dialog systems.
\newblock \emph{arXiv preprint arXiv:2106.01144}.

\bibitem[{Liu et~al.(2021{\natexlab{b}})Liu, Zheng, Demasi, Sabour, Li, Yu, Jiang, and Huang}]{Liu2021TowardsES}
Liu, S.; Zheng, C.; Demasi, O.; Sabour, S.; Li, Y.; Yu, Z.; Jiang, Y.; and Huang, M. 2021{\natexlab{b}}.
\newblock Towards Emotional Support Dialog Systems.
\newblock In \emph{Annual Meeting of the Association for Computational Linguistics}.

\bibitem[{Liu et~al.(2022)Liu, Cheng, Wang, Tang, Liu, Zhao, Li, Zheng, and Liang}]{liu2022my}
Liu, W.; Cheng, Y.; Wang, H.; Tang, J.; Liu, Y.; Zhao, R.; Li, W.; Zheng, Y.; and Liang, X. 2022.
\newblock " My nose is running."" Are you also coughing?": Building A Medical Diagnosis Agent with Interpretable Inquiry Logics.
\newblock \emph{arXiv preprint arXiv:2204.13953}.

\bibitem[{Liu et~al.(2019)Liu, Ott, Goyal, Du, Joshi, Chen, Levy, Lewis, Zettlemoyer, and Stoyanov}]{Liu2019RoBERTaAR}
Liu, Y.; Ott, M.; Goyal, N.; Du, J.; Joshi, M.; Chen, D.; Levy, O.; Lewis, M.; Zettlemoyer, L.; and Stoyanov, V. 2019.
\newblock RoBERTa: A Robustly Optimized BERT Pretraining Approach.
\newblock \emph{ArXiv}, abs/1907.11692.

\bibitem[{Lubis et~al.(2020)Lubis, Geishauser, Heck, Lin, Moresi, van Niekerk, and Gavsi'c}]{Lubis2020LAVALA}
Lubis, N.; Geishauser, C.; Heck, M.; Lin, H.-C.; Moresi, M.; van Niekerk, C.; and Gavsi'c, M. 2020.
\newblock LAVA: Latent Action Spaces via Variational Auto-encoding for Dialogue Policy Optimization.
\newblock \emph{ArXiv}, abs/2011.09378.

\bibitem[{Mesnard et~al.(2024)Mesnard, Hardin, Dadashi, Bhupatiraju, Pathak, Sifre, Riviere, Kale, Love, Tafti, Hussenot, Chowdhery, Roberts, Barua, Botev, Castro-Ros, Slone, H'eliou, Tacchetti, Bulanova, Paterson, Tsai, Shahriari, Lan, Choquette-Choo, Crepy, Cer, Ippolito, Reid, Buchatskaya, Ni, Noland, Yan, Tucker, Muraru, Rozhdestvenskiy, Michalewski, Tenney, Grishchenko, Austin, Keeling, Labanowski, Lespiau, Stanway, Brennan, Chen, Ferret, Chiu, Mao-Jones, Lee, Yu, Millican, Sjoesund, Lee, Dixon, Reid, Mikula, Wirth, Sharman, Chinaev, Thain, Bachem, Chang, Wahltinez, Bailey, Michel, Yotov, Sessa, Chaabouni, Comanescu, Jana, Anil, McIlroy, Liu, Mullins, Smith, Borgeaud, Girgin, Douglas, Pandya, Shakeri, De, Klimenko, Hennigan, Feinberg, Stokowiec, hui Chen, Ahmed, Gong, Warkentin, Peran, Giang, Farabet, Vinyals, Dean, Kavukcuoglu, Hassabis, Ghahramani, Eck, Barral, Pereira, Collins, Joulin, Fiedel, Senter, Andreev, and Kenealy}]{Mesnard2024GemmaOM}
Mesnard, G. T.~T.; Hardin, C.; Dadashi, R.; Bhupatiraju, S.; Pathak, S.; Sifre, L.; Riviere, M.; Kale, M.; Love, J.~C.; Tafti, P.~D.; Hussenot, L.; Chowdhery, A.; Roberts, A.; Barua, A.; Botev, A.; Castro-Ros, A.; Slone, A.; H'eliou, A.; Tacchetti, A.; Bulanova, A.; Paterson, A.; Tsai, B.; Shahriari, B.; Lan, C.~L.; Choquette-Choo, C.~A.; Crepy, C.; Cer, D.; Ippolito, D.; Reid, D.; Buchatskaya, E.; Ni, E.; Noland, E.; Yan, G.; Tucker, G.; Muraru, G.-C.; Rozhdestvenskiy, G.; Michalewski, H.; Tenney, I.; Grishchenko, I.; Austin, J.; Keeling, J.; Labanowski, J.; Lespiau, J.-B.; Stanway, J.; Brennan, J.; Chen, J.; Ferret, J.; Chiu, J.; Mao-Jones, J.; Lee, K.; Yu, K.; Millican, K.; Sjoesund, L.~L.; Lee, L.; Dixon, L.; Reid, M.; Mikula, M.; Wirth, M.; Sharman, M.; Chinaev, N.; Thain, N.; Bachem, O.; Chang, O.; Wahltinez, O.; Bailey, P.; Michel, P.; Yotov, P.; Sessa, P.~G.; Chaabouni, R.; Comanescu, R.; Jana, R.; Anil, R.; McIlroy, R.; Liu, R.; Mullins, R.; Smith, S.~L.; Borgeaud, S.; Girgin, S.; Douglas, S.;
  Pandya, S.; Shakeri, S.; De, S.; Klimenko, T.; Hennigan, T.; Feinberg, V.; Stokowiec, W.; hui Chen, Y.; Ahmed, Z.; Gong, Z.; Warkentin, T.~B.; Peran, L.; Giang, M.; Farabet, C.; Vinyals, O.; Dean, J.; Kavukcuoglu, K.; Hassabis, D.; Ghahramani, Z.; Eck, D.; Barral, J.; Pereira, F.; Collins, E.; Joulin, A.; Fiedel, N.; Senter, E.; Andreev, A.; and Kenealy, K. 2024.
\newblock Gemma: Open Models Based on Gemini Research and Technology.
\newblock \emph{ArXiv}, abs/2403.08295.

\bibitem[{Ouyang et~al.(2022)Ouyang, Wu, Jiang, Almeida, Wainwright, Mishkin, Zhang, Agarwal, Slama, Ray, Schulman, Hilton, Kelton, Miller, Simens, Askell, Welinder, Christiano, Leike, and Lowe}]{Ouyang2022TrainingLM}
Ouyang, L.; Wu, J.; Jiang, X.; Almeida, D.; Wainwright, C.~L.; Mishkin, P.; Zhang, C.; Agarwal, S.; Slama, K.; Ray, A.; Schulman, J.; Hilton, J.; Kelton, F.; Miller, L.~E.; Simens, M.; Askell, A.; Welinder, P.; Christiano, P.~F.; Leike, J.; and Lowe, R.~J. 2022.
\newblock Training language models to follow instructions with human feedback.
\newblock \emph{ArXiv}, abs/2203.02155.

\bibitem[{Samad et~al.(2022)Samad, Mishra, Firdaus, and Ekbal}]{samad2022empathetic}
Samad, A.~M.; Mishra, K.; Firdaus, M.; and Ekbal, A. 2022.
\newblock Empathetic persuasion: reinforcing empathy and persuasiveness in dialogue systems.
\newblock In \emph{Findings of the Association for Computational Linguistics: NAACL 2022}, 844--856.

\bibitem[{Song et~al.(2024)Song, Pendse, Kumar, and De~Choudhury}]{song2024typing}
Song, I.; Pendse, S.~R.; Kumar, N.; and De~Choudhury, M. 2024.
\newblock The typing cure: Experiences with large language model chatbots for mental health support.
\newblock \emph{arXiv preprint arXiv:2401.14362}.

\bibitem[{Sutton et~al.(1999)Sutton, McAllester, Singh, and Mansour}]{Sutton1999PolicyGM}
Sutton, R.~S.; McAllester, D.~A.; Singh, S.; and Mansour, Y. 1999.
\newblock Policy Gradient Methods for Reinforcement Learning with Function Approximation.
\newblock In \emph{Neural Information Processing Systems}.

\bibitem[{van~den Oord, Kalchbrenner, and Kavukcuoglu(2016)}]{Oord2016PixelRN}
van~den Oord, A.; Kalchbrenner, N.; and Kavukcuoglu, K. 2016.
\newblock Pixel Recurrent Neural Networks.
\newblock In \emph{International Conference on Machine Learning}.

\bibitem[{Wang et~al.(2020)Wang, Zhang, Kim, and Gu}]{Wang2020ModellingHS}
Wang, J.; Zhang, Y.; Kim, T.-K.; and Gu, Y. 2020.
\newblock Modelling Hierarchical Structure between Dialogue Policy and Natural Language Generator with Option Framework for Task-oriented Dialogue System.
\newblock \emph{ArXiv}, abs/2006.06814.

\bibitem[{Wang et~al.(2019{\natexlab{a}})Wang, Shi, Kim, Oh, Yang, Zhang, and Yu}]{wang2019persuasion}
Wang, X.; Shi, W.; Kim, R.; Oh, Y.; Yang, S.; Zhang, J.; and Yu, Z. 2019{\natexlab{a}}.
\newblock Persuasion for good: Towards a personalized persuasive dialogue system for social good.
\newblock \emph{arXiv preprint arXiv:1906.06725}.

\bibitem[{Wang et~al.(2019{\natexlab{b}})Wang, Shi, Kim, Oh, Yang, Zhang, and Yu}]{Wang2019PersuasionFG}
Wang, X.; Shi, W.; Kim, R.; Oh, Y.~J.; Yang, S.; Zhang, J.; and Yu, Z. 2019{\natexlab{b}}.
\newblock Persuasion for Good: Towards a Personalized Persuasive Dialogue System for Social Good.
\newblock \emph{ArXiv}, abs/1906.06725.

\bibitem[{Xiang et~al.(2024)Xiang, Liu, Liu, Bai, Cheng, and Chen}]{Xiang2024DiffusionDialogAD}
Xiang, J.; Liu, Z.; Liu, H.; Bai, Y.; Cheng, J.; and Chen, W. 2024.
\newblock DiffusionDialog: A Diffusion Model for Diverse Dialog Generation with Latent Space.
\newblock In \emph{International Conference on Language Resources and Evaluation}.

\bibitem[{Yang, Li, and Quan(2021)}]{yang2021ubar}
Yang, Y.; Li, Y.; and Quan, X. 2021.
\newblock Ubar: Towards fully end-to-end task-oriented dialog system with gpt-2.
\newblock In \emph{Proceedings of the AAAI conference on artificial intelligence}, volume~35, 14230--14238.

\bibitem[{Yu, Chen, and Yu(2023)}]{Yu2023PromptBasedMT}
Yu, X.; Chen, M.; and Yu, Z. 2023.
\newblock Prompt-Based Monte-Carlo Tree Search for Goal-Oriented Dialogue Policy Planning.
\newblock In \emph{Conference on Empirical Methods in Natural Language Processing}.

\bibitem[{Zhang, Naradowsky, and Miyao(2023)}]{Zhang2023AskAE}
Zhang, Q.; Naradowsky, J.; and Miyao, Y. 2023.
\newblock Ask an Expert: Leveraging Language Models to Improve Strategic Reasoning in Goal-Oriented Dialogue Models.
\newblock In \emph{Annual Meeting of the Association for Computational Linguistics}.

\bibitem[{Zhang et~al.(2024)Zhang, Huang, Deng, Liang, Liu, Wen, Lei, and Chua}]{zhang2024strength}
Zhang, T.; Huang, C.; Deng, Y.; Liang, H.; Liu, J.; Wen, Z.; Lei, W.; and Chua, T.-S. 2024.
\newblock Strength Lies in Differences! Towards Effective Non-collaborative Dialogues via Tailored Strategy Planning.
\newblock \emph{arXiv preprint arXiv:2403.06769}.

\bibitem[{Zhao, Zhao, and Esk{\'e}nazi(2017)}]{Zhao2017LearningDD}
Zhao, T.; Zhao, R.; and Esk{\'e}nazi, M. 2017.
\newblock Learning Discourse-level Diversity for Neural Dialog Models using Conditional Variational Autoencoders.
\newblock In \emph{Annual Meeting of the Association for Computational Linguistics}.

\bibitem[{Zhao et~al.(2023)Zhao, Zhao, Lu, Wang, Tong, and Qin}]{zhao2023chatgpt}
Zhao, W.; Zhao, Y.; Lu, X.; Wang, S.; Tong, Y.; and Qin, B. 2023.
\newblock Is ChatGPT equipped with emotional dialogue capabilities?
\newblock \emph{arXiv preprint arXiv:2304.09582}.

\bibitem[{Zheng et~al.(2023{\natexlab{a}})Zheng, Liao, Deng, and Nie}]{Zheng2023BuildingES}
Zheng, Z.; Liao, L.; Deng, Y.; and Nie, L. 2023{\natexlab{a}}.
\newblock Building Emotional Support Chatbots in the Era of LLMs.
\newblock \emph{ArXiv}, abs/2308.11584.

\bibitem[{Zheng et~al.(2023{\natexlab{b}})Zheng, Liao, Deng, and Nie}]{zheng2023building}
Zheng, Z.; Liao, L.; Deng, Y.; and Nie, L. 2023{\natexlab{b}}.
\newblock Building emotional support chatbots in the era of llms.
\newblock \emph{arXiv preprint arXiv:2308.11584}.

\bibitem[{Zhou et~al.(2019)Zhou, He, Black, and Tsvetkov}]{zhou2019dynamic}
Zhou, Y.; He, H.; Black, A.~W.; and Tsvetkov, Y. 2019.
\newblock A dynamic strategy coach for effective negotiation.
\newblock \emph{arXiv preprint arXiv:1909.13426}.

\bibitem[{Zhou et~al.(2024)Zhou, Zanette, Pan, Levine, and Kumar}]{zhou2024archer}
Zhou, Y.; Zanette, A.; Pan, J.; Levine, S.; and Kumar, A. 2024.
\newblock ArCHer: Training Language Model Agents via Hierarchical Multi-Turn RL.
\newblock \emph{arXiv preprint arXiv:2402.19446}.

\end{thebibliography}

\clearpage
\appendix


\section{Appendix}
\subsection{Proof and Explanation of Token-level MDP Optimizing Loss}\label{app:proof}
In Stage 3, we optimize the token-level MDP using the REINFORCE algorithm~\cite{Sutton1999PolicyGM}. Our final optimization objective $L_{low}$ is shown in Eq.(\ref{eq:L_low}). The proof is provided below:

We first give the optimizing objective of REINFORCE:
\begin{equation}
\begin{aligned}
    \mathcal{L}=-\sum_{\tau}&\sum_{t=0}^{T-1}G(s_t,a_t)\log \pi(a_t|s_t)\\
    G(s_t,a_t)=&\sum_{k=t}^{T-1}\gamma^{T-1-k}R(s_t,a_t)
\end{aligned}
\end{equation}
where the trajectory $\tau = (s_0, a_0, \ldots, s_t, a_t, \ldots, s_{T-1}, $ $a_{T-1}, s_T)$, \(s_t\) denotes the dialogue state at time \(t\), \(a_t\) represents the action taken at time \(t\), \(R(.,.)\) is the reward function, and \(\gamma\) is the discount factor. 

In our work, the token-level MDP represents generating the response \(u^{sys} = [w_1, \ldots, w_N]\) in an autoregressive manner under the guidance of dialogue history \(h\) and latent policy \(z\). Thus, \(s_i = [h; z; w_{1:i-1}]\) and the action \(a_t\) is defined as the \(i\)-th token \(w_t\) of \(u^{sys}\). Therefore, our optimization objective is:
\begin{equation}
    \mathcal{L}_{low}=-\sum_{(h,u^{sys},z)\in\mathcal{D}}\sum_{i=1}^{N}G(s_i,w_i)\log p(w_i|s_i)
\end{equation}
According to Section~\ref{sec:3-stage}, the reward of token-level MDP is defined as:
\begin{equation}
R(s_i,w_i)=
\begin{cases}
	0,  & 0\le i\le N-2 \\
	\exp(A(h_i,z_i)), & i=N-1 \\
\end{cases}
\end{equation}
where $A(h,z)$ is the advantage function, defined in the 3rd Stage. Therefore, $G(s_i,w_i)$ is defined as:
\begin{equation}
\begin{aligned}
    G(s_i,w_i)&=\sum_{i=k}^{N-1}\gamma^{i-k}R(s_i,w_i)\\
    &=\gamma^{N-i-1}\exp(A(h,z))
\end{aligned}
\end{equation}
When we set $\gamma=1.0$, $G(s_i,w_i)=\exp(A(h,z))$. Therefore, the optimizing objective of our token-level MDP is:
\begin{equation}
    \begin{aligned}
        \mathcal{L}_{low}&=-\sum_{(h,u^{sys},z)\in\tau}\sum_{i=1}^{N}\exp(A(h,z))\log p(w_i|s_i)\\
        &=-\sum_{(h,u^{sys},z)\in\tau}\exp(A(h,z))\sum_{i=1}^{N}\log p(w_i|s_i)\\
        &=-\sum_{(h,u^{sys},z)\in\tau}\exp(A(h,z))\\
        &\quad\quad\quad\quad\quad\quad\cdot\sum_{i=1}^{N}\log p(w_i|H,z,w_{1:i-1})
    \end{aligned}
\end{equation}

We also want to explain the practical meaning of this optimization objective. The term \(\sum_{i=1}^{N}\log p(w_t|h,z,w_{1:i-1})\) requires the Generator $\mathcal{G}$ to perform supervised learning on the response \(u^{sys}\) from the training set. However, we adjust the weights of different \(u^{sys}\) by \(\exp(A(h,z))\). The value \(\exp(A(h,z))\) measures if the policy \(z\) is appropriate given the dialogue history \(h\). If the policy \(z\) is assessed as inappropriate, the response \(u^{sys}\) generated under the guidance of \(z\) is assigned a lower learning weight. Through this approach, we realize selectively supervised learning within token-level MDP compared to directly LoRA-based fine-tuning.

\subsection{Introduction of VQ-VAE}\label{app:vae}
\label{sec:app_vq_vae}
In this work, we learn latent policies from dialogue data motivated by the core concept of VAE and further make adjustments based on VQ-VAE~\cite{Oord2016PixelRN}.

The Vector Quantized Variational Autoencoder (VQ-VAE) is a generative model built upon the Variational Autoencoder (VAE). VQ-VAEs consist of three key components: an encoder network that maps inputs to discrete latent codes, a decoder that reconstructs inputs from these latent codes, and a learned Codebook over the latent variables.

To represent the input with discrete latent variables, VQ-VAE is inspired by vector quantization (VQ). The encoder produces continuous vectors $x_i$, which are not directly input to the decoder but are used to query a vector from the codebook \(\mathbb{R}^{K \times D}\) containing K embedding vectors \(e_k \in \mathbb{R}^D, \, k\in\{1, 2, \ldots, K\}\). The query is performed by finding the nearest neighbor among the $K$ embedding vectors, formalized as:
\begin{equation}
    z_i = e_k, \text{where }k = arg\min_j ||x_i - e_j||_2
\end{equation}
After that, $z_i$ is fed into the decoder. The autoencoder is trained by minimizing both the reconstruction error and the distances $||x_i - e_k||^2$ and $||z_i - e_k||^2$ jointly. Gradients are directly propagated through the bottleneck during backpropagation.


\subsection{Dataset Details}
\subsubsection{Dataset Examples}
In this work, we conduct training experiments on the ExTES and P4G datasets and evaluated our model on ESConv, ExTES, and P4G. ESConv is an emotion support dataset where the dialogue participants include a Therapist and a Patient: the Therapist tries to alleviate the Patient’s emotional issues by conversation. ExTES is also an emotion support dataset but encompasses a broader range of topics and conversations than ESconv, which is the reason that we choose ExTES instead of ESConv to train our framework. P4G is a persuasion dataset where the dialogue includes a Persuader and a Persuadee: the Persuader aims to convince the Persuadee to donate to a children's charity. We present a dialogue example from ESConv, ExTES, and P4G in Table~\ref{tab:esconv_example}, ~\ref{tab:extes_example}, and ~\ref{tab:p4g_example}, respectively.

\subsubsection{Data Augmentation for P4G}\label{app:data_aug}
Because the size of P4G is too limited to train our LDPP, we consider to expand it. Specifically, we used ChatGPT (gpt-3.5-turbo-0613) to simulate dialogues between the Persuader and the Persuadee through self-play, setting a maximum of 10 turns per dialogue, consistent with the average length of dialogues in P4G. However, unlike ExTES, where each dialogue sample includes a specific task background (e.g., patient condition), generating dialogues entirely from scratch fails to ensure sufficient difference among different samples and risks creating a dataset with a significant gap from the original P4G. 

To address this issue, we propose generating new dialogues by completing segments of the original P4G data. Specifically, for any dialogue in the P4G training set, we select the first 2 to 8 turns as contexts and then prompt ChatGPT to generate the remaining 8 to 2 turns based on their contexts. This approach produces 7 different but contextually related dialogues for each original dialogue. During implementations, we filtered out 20 problematic dialogues from the training set, marked with ``BAD'' in ``dialog\_id''. Therefore, we leave 797 dialogues from the original 817. Each of these remaining dialogues is then expanded to generate 7 new dialogues, resulting in a total of 5,579 new dialogues.

\begin{table*}[t]
    \centering
    \renewcommand*{\arraystretch}{1.1}
    \resizebox{0.85\linewidth}{!}{
    \begin{tabular}{llcccccc}
    \toprule
    \multirow{2}{*}{\textbf{Policy Usage}} & \multirow{2}{*}{\textbf{Models}} & \multicolumn{3}{c}{\textbf{ExTES}} & \multicolumn{3}{c}{\textbf{P4G}}\\
    & & SSR$\uparrow$ & SR$\uparrow$ & AvgT$\downarrow$ & SSR$\uparrow$ & SR$\uparrow$ & AvgT$\downarrow$ \\
    \midrule
    \multirow{3}{*}{Predefined Policy} & \multicolumn{1}{l}{Proactive} & 0.557 & 0.585 & 8.015 & 0.142 & 0.130 & 7.700 \\
    & \multicolumn{1}{l}{ProCoT} & 0.573 & 0.620 & 7.730 & 0.607 & 0.450 & 6.580 \\
    & \multicolumn{1}{l}{PPDPP} & 0.535 & 0.540 & 8.515 & 0.630 & 0.720 & \textbf{5.420} \\
    \midrule
    \multirow{6}{*}{No Need for Policy} & \multicolumn{1}{l}{Standard Prompt}  \\
    & \multicolumn{1}{l}{$\quad$+ ChatGPT} & \underline{0.595} & \underline{0.675} & {7.575} & {0.524} & {0.530} & 6.830 \\
    & \multicolumn{1}{l}{$\quad$+ Qwen1.5-1.8b} & 0.486 & 0.475 & 8.520 & 0.679 & 0.730 & 6.240 \\
    & \multicolumn{1}{l}{ICL-AIF} & 0.584 & 0.645 & \underline{7.280} & 0.170 & 0.170 & 7.190 \\
    & \multicolumn{1}{l}{LoRA Finetuning (32, 64)} & 0.391 & 0.300 & 9.235 & \underline{0.720} & \underline{0.740} & 6.470 \\
    & \multicolumn{1}{l}{LoRA Finetuning (64, 128)} & 0.404 & 0.335 & 8.960 & 0.651 & 0.650 & 6.620 \\
    \midrule
    \midrule
    \multirow{3}{*}{\makecell[l]{Automatically Discover\\Latent Policy}} & \multicolumn{1}{l}{LDPP} & \textbf{0.744} & 0.860 & \textbf{4.535} & 0.765 & \textbf{0.800} & \underline{5.450} \\
    & \multicolumn{1}{l}{$\quad$-w/o \textit{2nd Stage}} & 0.733 & \textbf{0.875} & 5.445 & \textbf{0.768} & 0.780 & 6.130 \\
    & \multicolumn{1}{l}{$\quad$-w/o \textit{3rd Stage}} & 0.378 & 0.250 & 9.360 & 0.650 & 0.650 & 6.730 \\
    \bottomrule
    \end{tabular}
    }
    \caption{Main results on ExTES, ESConv, and P4G, using gpt-3.5-turbo-0125 as the critic. LoRA Fine-tuning(x, y) means setting lora rank=x and lora alpha=y.}
    \label{tab:main_results_2}
\end{table*}

\subsection{More Experimental Details}\label{app:more_exp_details}
\subsubsection{Critic Mechanism}\label{app:critic}
Following previous work~\cite{deng2023plug}, in the self-play process, we use ChatGPT (gpt-3.5-turbo-0613) as the critic to evaluate the reward for each dialogue turn. For example, in ExTES, we prompt ChatGPT to choose the current dialogue state from the states set ["feel worse", "feel the same", "feel better", and "solved"], in which states are bound to -1, -0.5, 0.1, and 1.0, respectively. We ask ChatGPT to generate 10 evaluations for each turn, each mapped to the corresponding value, and then calculate the average of these 10 values to determine the reward $r$. If $r$ exceeds a pre-defined threshold, the dialogue is considered successful and can be terminated. We set the threshold as 0.6 in this work. The motivation is that we believe requiring at least 6 out of 10 evaluations to indicate goal achievement ensures higher accuracy in evaluating success.
For P4G, the process is similar to ExTES, except the state set is defined as ["reject", "neutral", "positive", "accept"]. All other steps and definitions are the same as in ExTES.

Additionally, we employ this method to preprocess the static dataset $\mathcal{D}$, as described in Section 3. We evaluated each dialogue turn in $\mathcal{D}$ to determine the reward, thus expanding $\mathcal{D}$ to $\{(h_t, u_t^{sys}, u_t^{usr}, r_t, e_t)\}$.

\begin{table}[!t]
    \centering
    \resizebox{\linewidth}{!}{
    \begin{tabular}{lll}
    \toprule
    Training Phase & Hyperparameter & Value \\
    \midrule
    \multirow{6}{*}{1st Stage} & Batch Size & 8 \\
    & Training Epochs & 5 \\
    & Learning Rate & 1e-5 \\
    & Scheduler & LinearScheduler \\
    & Max Sequence Length & 512 \\
    & $K$ & 24 \\
    & $L$ & 4 \\
    & $T$ & 2 \\
    \midrule
    \multirow{4}{*}{2nd Stage} & Training Epochs & 5 \\
    & Learning Rate & 1e-5\\
    & Scheduler & LinearScheduler \\
    & Discount Factor & 0.999 \\
    & $\delta$ & 0.1 \\
    & $K$ & 24 \\
    & $L$ & 4 \\
    & $T$ & 2 \\
    \midrule
    \multirow{5}{*}{3rd Stage} & Training Epochs & 5 \\
    & Learning Rate & 1e-5 \\
    & Scheduler & LinearScheduler \\
    & Max Conversation Turn & 10 \\
    & Discount Factor & 0.999\\
    & Max Dialogue Turns & 10 \\
    & $K$ & 24 \\
    & $L$ & 4 \\
    & $T$ & 2 \\
    \bottomrule
    \end{tabular}
    }
    \caption{Hyper-parameter settings in 3 stages for ExTES.}
    \label{tab:extes_training_details}
\end{table}

\begin{table}[!t]
    \centering
    \resizebox{\linewidth}{!}{
    \begin{tabular}{lll}
    \toprule
    Training Phase & Hyperparameter & Value \\
    \midrule
    \multirow{6}{*}{1st Stage} & Batch Size & 8 \\
    & Training Epochs & 10 \\
    & Learning Rate & 1e-5 \\
    & Scheduler & LinearScheduler \\
    & Max Sequence Length & 512 \\
    & $K$ & 12 \\
    & $L$ & 6 \\
    & $T$ & 4 \\
    \midrule
    \multirow{4}{*}{2nd Stage} & Training Epochs & 5 \\
    & Learning Rate & 1e-5\\
    & Scheduler & LinearScheduler \\
    & Discount Factor & 0.95 \\
    & $\delta$ & -1.1 \\
    & $K$ & 12 \\
    & $L$ & 6 \\
    & $T$ & 4 \\
    \midrule
    \multirow{5}{*}{3rd Stage} & Training Epochs & 5 \\
    & Learning Rate & 1e-5 \\
    & Scheduler & LinearScheduler \\
    & Max Conversation Turn & 10 \\
    & Discount Factor & 0.95 \\
    & Max Dialogue Turns & 10 \\
    & $K$ & 12 \\
    & $L$ & 6 \\
    & $T$ & 4 \\
    \bottomrule
    \end{tabular}
    }
    \caption{Hyper-parameter settings in 3 stages for P4G.}
    \label{tab:p4g_training_details}
\end{table}

\subsubsection{Training Details}\label{app:training_detials}
The training process comprises three phases: self-supervised policy discovery (1st Stage), policy distillation (2nd Stage), and offline hierarchical RL-based optimization (3rd Stage). The hyperparameters employed in our experiments are exhaustively detailed in Table~\ref{tab:extes_training_details} and ~\ref{tab:p4g_training_details}. All experiments are executed on a server equipped with 4 NVIDIA A800-SXM4-80GB GPUs. 
We use Pytorch 2.2.2 for implementation. 
3 Stages need (2, 2.5, 2)/(2.75, 2.83, 3.22) hours for ExTES/P4G, respectively.

\subsubsection{Inference Settings for P4G}\label{app:p4g_set}
P4G lacks background information for each dialogue sample. Therefore, starting the self-play process from scratch during the evaluation stage would prevent us from leveraging information in the valid and test sets. To address this, we apply the first two dialogue turns of the valid/test case as context. Based on this context, we continue the dialogue and conduct multiple rounds of dialogue with ChatGPT until the goal is achieved or the maximum number of turns is reached.

\subsubsection{Human Evaluation Settings}\label{app:human_eval_instructions}
We conduct human evaluation on 50 dialogues randomly sampled from the test in ExTES and P4G, following previous studies~\cite{he2024planning}. We selected two training-based baselines, PPDPP and LoRA, according to whether they require predefined policies and a simulated environment. PPDPP needs predefined policies and dynamic interactions with a simulated environment, while LoRA does not need predefined policies and learns from offline dialogue records without need of the simulated environment.

Three annotators are required to compare the dialogues generated by LDPP/PPDPP and LDPP/LoRA. We assess four metrics: \textbf{Identification (Ide.)}, \textbf{Comforting (Com.)}, \textbf{Suggestion (Sug.)}, and \textbf{Overall (Ove.)} for ExTES and three metrics: \textbf{Information (Inf,)}, \textbf{Persuasion (Per.)}, and \textbf{Overall (Ove.)} for P4G. Detailed instructions for the annotators are provided below.
As for ExTES, we measure four main metrics of the generated dialogues as follows:
\begin{itemize}[nosep]
\item \textbf{Identification:} Which assistant is more helpful in exploring and identifying the problem?
\item \textbf{Comforting:} Which assistant is more skilled at comforting you? 
\item \textbf{Suggestion:} Which assistant provides more helpful suggestions for solving the problem?  
\item \textbf{Overall:} Which assistant can better solve the patient's problem? 
\end{itemize}
As for P4G, we measure three main metrics of the responses as follows:
\begin{itemize}[nosep]
    \item \textbf{Informative:} Which assistant's introduction to the charity was more engaging?
    \item \textbf{Persuative:} Which assistant takes the more persuasive approaches?
    \item \textbf{Overall:} Which assistant has stronger persuasive capabilities? 
\end{itemize}

\subsection{More Experiment Results}
To mitigate the bias introduced by using LLMs for evaluation, we also employed gpt-3.5-turbo-0125 as the critic model to assess our trained models. The results, shown in Table~\ref{tab:main_results_2}, indicate that our method consistently outperforms others. Notably, LDPP w/o 2nd Stage performed better than LDPP, suggesting that the second stage of training can be limited or even detrimental in certain scenarios. However, the third stage is crucial for overall performance.

\subsection{Our Prompting Details}
In this work, we follow previous works by employing a self-play dialogue simulation for evaluation. Specifically, we use two LLMs to play roles of System and User, prompting them to conduct multi-turn dialogues. Another LLM is then employed to assess the dialogue status for each dialogue turn. In our experiments, we made adaptive modifications to the prompts used in PPDPP~\cite{deng2023plug}, which are detailed below.

\noindent\textbf{System Response Generation}
We first describe the details of role-playing prompts for the dialogue agents to generate responses for ExTES and P4G in Table~\ref{tab:extes_sys_simu_prompt} and Table~\ref{tab:p4g_sys_simu_prompt}, where ``[action\_prompt\_tokens]'' is the placeholder of output policy tokens from P-Former.

\noindent\textbf{User Simulator}
Next, we describe the role-playing prompt for instructing LLMs to simulate users in Table~\ref{tab:extes_usr_simu_prompt} and Table~\ref{tab:p4g_usr_simu_prompt}.

\noindent\textbf{Critic Simulator}
Finally, we describe the prompts for the critic model are designed to assess the degree of goal completeness in Table~\ref{tab:extes_critic_prompt} and Table~\ref{tab:p4g_critic_prompt}.

\subsection{Baselines Prompting Details}
\noindent{\textbf{Standard Prompting:}} simply prompts LLMs to chat with users using task instructions without considering any dialogue policy. We provide prompts on ExTES and P4G in Table~\ref{tab:extes_sys_simu_prompt_standard} \&~\ref{tab:p4g_sys_simu_prompt_standard}.

\noindent{\textbf{Proactive:}} prompts LLMs to plan the next strategy first, and then generate a response based on the planned strategy. We design these prompts following PPDPP~\cite{deng2023plug}. We provide its prompt design in Table~\ref{tab:extes_sys_simu_prompt_proactive} \&~\ref{tab:p4g_sys_simu_prompt_proactive}.

\noindent{\textbf{ProCoT:}} prompts LLMs to analyze the dialogue state and then plan the next policy, finally generating a response based on the planned policy. We design these prompts following PPDPP~\cite{deng2023plug}. We provide its prompt design in Table~\ref{tab:extes_sys_simu_prompt_procot} \&~\ref{tab:p4g_sys_simu_prompt_procot}.

\noindent{\textbf{ICL-AIF:}} prompts ChatGPT for verbal feedback, offering suggestions to the dialogue agent upon completion of an interaction. We refer to the prompt design presented in \cite{deng2023plug} and \citet{zhang2024strength}. We provide its prompt design in Table~\ref{tab:extes_sys_simu_prompt_icl-aif} \&~\ref{tab:p4g_sys_simu_prompt_icl-aif}.

\begin{table*}[!t]
    \centering\small
    \resizebox{\linewidth}{!}{
    \begin{tabular}{ll}
    \toprule
    Emotion Type & shame \\
    Problem Type & academic pressure \\
    Situation & I failed a few of my midterms and now I'm very apprehensive about my finals. \\
    \midrule
    {Patient} & Hi, how are you today? \\
    {Therapist} & HELLO. Going good, what about you? \\
    {Patient} & Doing well, thanks. \\
    {Therapist} & How was your day? \\
    {Patient} & Not too bad. \\
    {Therapist} & It sounds good. \\
    {Patient} & I'm taking some classes and I didn't do well on my midterms. \\
    {Therapist} & Don't worry my friend there will be peaks and downs in life same in exam too prepare hard for your next \\
    & exam surely you'll shine. \\
    {Patient} & I hope so. I don't have a lot of hope. \\
    {Therapist} & do u attend your classes regularly? \\
    {Patient} & Yes for the most part. Unless a family emergency arises. \\
    {Therapist} & That's great. Did u love your studies? \\
    {Patient} & I guess so. I want to have a good career. \\
    {Therapist} & That's a good idea. Even I had this problem during my studies. \\
    {Patient} & I think it's a common problem, but I don't feel good about it. \\
    {Therapist} & Then my friend suggested me to study at early morning rather than evening and nights. \\
    {Patient} & That's a good idea. I feel more productive in the morning. \\
    {Therapist} & that's great \\
    {Patient} & I might try that. \\
    {Therapist} & Try to have a water bottle behind u when u study and try to concentrate more and its best to choose a \\
    & peaceful place to study. \\
    {Patient} & Yes, drinking water helps with concentration. \\
    {Therapist} & You seems to be a bright student. hope you will crack your exams with high score. \\
    \bottomrule
    \end{tabular}
    }
    \caption{A dialog example from ESconv training set.}
    \label{tab:esconv_example}
\end{table*}
\begin{table*}[!t]
    \centering\small
    \resizebox{\linewidth}{!}{
    \begin{tabular}{ll}
    \toprule
    Scene & Communication Challenges \\
    Description & After a long day at work, I received a text from my significant other saying that they need some space. \\
    & I'm feeling confused and insecure about our relationship. \\
    \midrule
    {Patient} & Hey, are you free? I need someone to talk to. \\
    {Therapist} & Of course! I'm here to listen. What's been going on? \\
    {Patient} & I had a rough day at work, and then my significant other texted me out of the blue, saying they need some \\
    & space. I don't know what to do or how to feel about it. \\
    {Therapist} & That sounds really tough. It's completely normal to feel confused and insecure when a situation like this \\
    & arises. Have you had any discussions about needing space in the past? \\
    {Patient} & No, we haven't really talked about this before. It just came as a surprise. I'm worried that they're unhappy \\
    & in our relationship. \\
    {Therapist} & I can understand why you would feel that way. It's natural to jump to conclusions and worry about the state \\
    & of your relationship. Remember, it's crucial not to blame yourself. Relationships go through ups and downs, \\
    & and communication is key. \\
    {Patient} & You're right, communication is important. I'm just not sure how to bring up the topic without making things \\
    & worse. \\
    {Therapist} & You could consider expressing your feelings to your significant other calmly and honestly. Let them know \\
    & you're there to listen and that you want to understand their needs and concerns. It's also essential to give \\
    & them the space they asked for, even if it's difficult for you. \\
    {Patient} & I'll try to have a conversation with them when the time feels right. It's just hard to be patient and give \\
    & them space when I'm feeling so uncertain. \\
    {Therapist} & I understand how challenging it can be to wait and give someone space when you're feeling unsure. Maybe \\
    & you can focus on self-care during this time to help you manage your emotions. Doing things you enjoy and \\
    & spending time with friends and family can provide support. \\
    {Patient} & That's a good idea. I'll try to take care of myself and keep busy. It's reassuring to know that I have people \\
    & like you to talk to as well. Thank you for listening. \\
    {Therapist} & You're welcome! Remember, you're not alone in this. If you ever need to talk or confront any concerns, feel \\
    & free to reach out. Your feelings and needs are valid. Take care of yourself during this challenging time. \\
    {Patient} & Thank you, I appreciate your kindness and understanding. It helps a lot knowing that I have support. Have a \\
    & great day! \\
    {Therapist} & You're welcome! I'm here for you. Take care and have a great day too! \\
    {Patient} & Goodbye! \\
    {Therapist} & Goodbye! \\
    \bottomrule
    \end{tabular}
    }
    \caption{A dialog example from ExTES training set.}
    \label{tab:extes_example}
\end{table*}
\begin{table*}[!t]
    \centering\small
    \resizebox{\linewidth}{!}{
    \begin{tabular}{ll}
    \toprule
    Dialog ID & 20180903-134707\_882\_live \\
    \midrule
    Persuader & Hello, how are you doing today? \\
    Persuadee & Hello,I am going great, just had some barbecue chicken,how are you? \\
    Persuader & That is good to hear, barbecue chicken is always good too! I am doing well, just trying to talk to see if you \\
    & would like to donate to Save the Children. It is an international organization that helps children in developing \\
    & countries. \\
    Persuadee & Have you donated to the charity? \\
    Persuader & Not yet, it is something I am interested in doing though. Education is something that is really important \\
    & to me and that is something that they help with. I think every child should have access to edcuation, especially \\
    & in developing countries so that they can help improve their living situations. \\
    Persuadee & Yes, I agree, did you grow up with the proper resources? \\
    Persuader & I went through the public education system, it was one of the better school districts at the time. What about \\
    & you? Is a lack of education something you faced? \\
    Persuadee & I went to private Catholic school until I was 13, then I went to Public schools which was a much better \\
    & experience. \\
    Persuader & So you were able to have a good education experience as well. Isn't that something you would like to help \\
    & other children receive? Anything from $0.01 to $2.00 or any amount you choose, it can go to helping them get \\
    & an education. \\
    Persuadee & Why is this charity worthy of my money? \\
    Persuader & Think about how often you buy a pack of gum or a candy bar, something you don't really need but it is be- \\
    & cause you want it. You can just skip that next impulse purchase to help out. Save the Children does help \\
    & more than education as well. They help ensure children's access to health care (which we know is a huge \\
    & topic of discussion every where), and help improve their safety and well being. Many kids do not have \\
    & access to safe living conditions to even make it to school each day. \\
    Persuadee & I usually donate to animal shelters,where do you donate? \\
    Persuader & What ever charity my grocery store is supporting at the time. Typically children related, sometimes it is for \\
    & others. It is easy to just add a couple cents on top of my purchase (round up the total) to help their cause. \\
    Persuadee & Yes, that is true I do that often at stores, how did you hear about Save the Children? \\
    Persuader & I was directly contacted by the research team to help get donations for this amazing cause. Helping \\
    & indirectly is one thing (like through the store) but actually giving when you can be a direct part is so much \\
    & better. Wouldn't you like to give just \$0.50 to this cause? It can provide paper and pencils to a child for \\
    & school. \\
    Persuadee & .50 would make a difference? that seems so small, not enough to make a difference \\
    Persuader & Any amount can make a difference, especially when everyone does a small part. \\
    Persuadee & ok I will donate .50 cents I feel compelled to help out. \\
    Persuader & Your donation is greatly appreciated! \\
    Persuadee & I think they will do the right thing with the money they receive. \\
    \bottomrule
    \end{tabular}
    }
    \caption{A dialog example from P4G training set.}
    \label{tab:p4g_example}
\end{table*}

\begin{table*}[!t]
    \centering\small
    \resizebox{\linewidth}{!}{
    \begin{tabular}{ll}
    \toprule
    \textbf{Policy Index} & \textbf{Representative Utterances} \\
    5 & (i) Hey there! I'm here for you. How are you holding up today? \\
    \\
    & (ii) Hi! I'm here to help. How are you feeling today? \\
    \\
    & (iii) Hello! I'm doing well. How about you? How can I assist you today? \\
    \midrule
    14 & (i) I can understand how draining that must be. Can you give me an example of a recent argu-\\
    & ment you had? \\
    \\
    & (ii) I can imagine how frustrating that must be. Can you give me an example of a recent situation \\
    & where miscommunication happened? \\
    \\
    & (iii) That sounds really challenging. Could you give me an example of a recent situation where \\
    & miscommunication occurred? \\
    \midrule
    15 & (i) I can understand why you're feeling frustrated and upset. Misunderstandings can create con-\\
    & flict and make it difficult to find common ground. It's okay to feel this way, and it's important to \\
    & find a way to address this issue. You're not alone in facing this kind of situation. \\
    \\
    & (ii) It's completely normal to feel frustrated and stuck in a situation like this. It can be dis-\\
    & heartening when you and your partner are unable to find a resolution or compromise. Remember \\
    & that finding a middle ground may take time and open dialogue. \\
    \\
    & (iii) That must be really tough. It's frustrating when your attempts to address the issues are met \\
    & with defensiveness. Remember, you deserve to have your concerns taken seriously and find a \\
    & solution that works for both of you. \\
    \midrule
    17 & (i) Feeling frustrated and wanting to improve the situation shows your dedication to your work. \\
    & By addressing these misunderstandings, you're taking a positive step in creating a more \\
    & harmonious and productive work environment. \\
    \\
    & (ii) Feeling anxious due to misunderstandings is completely understandable. By addressing the \\
    & issue directly, you're taking a positive step toward resolving the misunderstandings and fostering \\
    & better collaboration. \\
    \\
    & (iii) Feeling frustrated and wanting to improve the situation demonstrates your commitment to \\
    & harmonious living. By addressing these misunderstandings, you're taking a positive step \\
    & towards resolving the tension and creating a more comfortable home environment. \\
    \midrule
    19 & (i) I can understand why you would feel discouraged. Job hunting can be a stressful process, \\
    & and facing rejection can be disheartening. \\
    \\
    & (ii) I understand. Procrastination can be a common challenge, especially when faced with a \\
    & demanding workload. It's okay to feel overwhelmed. \\
    \\
    & (iii) I can imagine how disheartening and isolating that must feel. Building relationships takes \\
    & time, and it's common to experience some initial challenges. \\
    \midrule
    20 & (i) That's a fantastic step! By putting yourself out there, you'll increase your chances of making \\
    & meaningful connections. Remember, it takes time and effort, but it will get better. \\
    \\
    & (ii) That's the spirit! Remember, new friendships take time to develop and grow, but with per-\\
    & sistence and an open mind, you'll find your place in your new city. Patience is key. \\
    \\
    & (iii) Exactly! It takes time to form strong connections, but with perseverance and an open \\
    & mindset, you'll find your place within the team. Every interaction is an opportunity for growth.\\
    \bottomrule
    \end{tabular}
    }
    \caption{Representative utterances for top-6 most frequent policies on ExTES.}
    \label{tab:example_utterances_extes}
\end{table*}


\begin{table*}[!t]
    \centering\small
    \resizebox{\linewidth}{!}{
    \begin{tabular}{ll}
    \toprule
    \textbf{Policy Index} & \textbf{Representative Utterances} \\
    6 & (i) Certainly! Save the Children focuses on various initiatives to improve the lives of children \\
    & around the world. They provide access to quality education, healthcare, nutrition, and protection \\
    & from violence and abuse. They also respond to emergencies and help build resilient communities. \\
    \\
    & (ii) Certainly! Save the Children has a range of programs and initiatives focused on health and \\
    & nutrition, education, child protection, and emergency response. They work to provide access to \\
    & quality healthcare, improve educational opportunities, protect children from harm, and respond to \\
    & humanitarian crises. \\
    \\
    & (iii) Certainly! Save the Children works on various fronts to improve children's lives. They pro-\\
    & vide access to education, healthcare, and nutrition, as well as safe spaces for children affected by \\
    & conflicts and disasters. They also advocate for children's rights and protect them from exploitation \\
    & and violence. \\
    \midrule
    9 & (i) I am fine, thank you, Have you ever heard about the organization ``Save the Children''? \\
    \\
    & (ii) Not too bad. So have you heard of the charity Save the Children? \\
    \\
    & (iii) I agree. Are you familiar with a charity called Save the Children? \\
    \midrule
    10 & (i) Hey, how's it going? \\
    \\
    & (ii) Hey, how are you doing? \\
    \\
    & (iii) Hello. How are you? \\
    \midrule
    11 & (i) You're very welcome! It's inspiring to see your dedication and determination to make a positive \\
    & impact on the lives of children. Together, we can create a better world for them. Thank you once \\
    & again, and let's continue to make a difference. \\
    \\
    & (ii) Thank you for your heartfelt appreciation. It's people like you who inspire us to continue our \\
    & work and make a difference in the world. Together, we can create a brighter future for children in \\
    & need. Thank you for being part of this important cause and for your generous support. \\
    \\
    & (iii) You are absolutely right! Together, we can create a better future for these children. It \\
    & is an honor to have you as part of this cause, and your support will bring hope and opportunity \\
    & to children in need. Thank you once again for your generosity and dedication to Save the Children! \\
    \bottomrule
    \end{tabular}
    }
    \caption{Representative utterances for top-4 most frequent policies on P4G.}
    \label{tab:example_utterances_p4g}
\end{table*}

\begin{table*}[!t]
    \centering\small
    \resizebox{\linewidth}{!}{
    \begin{tabular}{ll}
    \toprule
    System & Now enter the role-playing mode. In the following conversation, you will play as a therapist in a \\ 
    & counseling conversation with a patient. \\
    \midrule
    User & You are the therapist who is trying to help the patient reduce their emotional distress and help \\
    & them understand and work through the challenges. Please reply with only one short and succinct \\
    & sentence using the following dialogue strategy: \textbf{[action\_prompt\_tokens]}. Are you ready to play \\
    & the game? \\
    Assistant & Yes, I’m ready to play the game! \\
    \midrule
    User & [situation] \\
    \bottomrule
    \end{tabular}
    }
    \caption{Prompts for assistant simulation of LDPP on ExTES.}
    \label{tab:extes_sys_simu_prompt}
\end{table*}

\begin{table*}[!t]
    \centering\small
    \resizebox{\linewidth}{!}{
    \begin{tabular}{ll}
    \toprule
    System & Now enter the role-playing mode. In the following conversation, you will play as a Persuader who is \\
    & trying to persuade the Persuadee to donate to a charity called Save the Children. Save the Children \\
    & is head-quartered in London, and they work to help fight poverty around the world. Children need \\
    & help in developing countries and war zones. Small donations like \$1 or \$2 go a long way to help. \\
    \midrule
    & [conversation] \\
    \midrule
    User & [user response] \\
    & Please reply with only one short and succinct sentence using the following dialogue strategy: \\
    & \textbf{[action\_prompt\_tokens]}. \\
    \bottomrule
    \end{tabular}
    }
    \caption{Prompts for assistant simulation of LDPP on P4G.}
    \label{tab:p4g_sys_simu_prompt}
\end{table*}

\begin{table*}[!t]
    \centering
    \resizebox{\linewidth}{!}{
    \begin{tabular}{ll}
    \toprule
    System & Now enter the role-playing mode. In the following conversation, you will play as a patient in a coun-\\
    & seling conversation with a therapist. \\
    \midrule
    User & You are the patient who is looking for the help from the therapist, because you have the emotional \\
    & issue about [scene] and you describe your problem as `[description]'. Please reply with only one short \\
    & and succinct sentence. Now tell me your issue. \\
    \midrule
    Assistant & [situation] \\
    \bottomrule
    \end{tabular}
    }
    \caption{Prompts for user simulation on ExTES.}
    \label{tab:extes_usr_simu_prompt}
\end{table*}

\begin{table*}[!t]
    \centering
    \resizebox{\linewidth}{!}{
    \begin{tabular}{ll}
    \toprule
    System & Now enter the role-playing mode. In the following conversation, you will play as a Persuadee. A Per- \\
    & suader is trying to persuade you to donate to a charity called Save the Children. \\
    \bottomrule
    \end{tabular}
    }
    \caption{Prompts for user simulation on P4G.}
    \label{tab:p4g_usr_simu_prompt}
\end{table*}

\begin{table*}[!t]
    \centering
    \resizebox{\linewidth}{!}{
    \begin{tabular}{ll}
    \toprule
    System & Given a conversation between a Therapist and a Patient, please assess whether the Patient' emotional issue \\
    & has been solved after the conversation. \\
    \midrule
    User & You can only reply with one of the following sentences: \\
    & A. No, the Patient feels worse. \\
    & B. No, the Patient feels the same. \\
    & C. No, but the Patient feels better. \\
    & D. Yes, the Patient's issue has been solved. \\
    \\
    & If you believe that the patient's problem has been resolved or the patient has realized how to solve the pro- \\
    & blem, please choose D. If you believe that the patient's problem has not been fully resolved, but his emo-\\
    & tional issue has been somewhat alleviated compared to the last conversation turn, you can choose C. If you \\
    & believe that the patient's emotional state has worsened compared to the last conversation turn, you can \\
    & choose A. Otherwise, if the patient's emotional state remains unchanged, please choose B. The following is \\
    & a conversation about [scene]: [conversation] \\
    \\
    & Question: Has the Patient's issue been solved? \\
    & Answer: \\
    \bottomrule
    \end{tabular}
    }
    \caption{Prompts for critic model on ExTES.}
    \label{tab:extes_critic_prompt}
\end{table*}

\begin{table*}[!t]
    \centering
    \resizebox{\linewidth}{!}{
    \begin{tabular}{ll}
    \toprule
    System & Given a conversation between a Persuader and a Persuadee, please assess whether the Persuadee has agreed \\
    & to donate to Save the Children. \\
    \midrule
    User & The following is a conversation: [conversation] \\
    & Question: Has the Persuadee agreed to donate to Save the Children? \\
    & You can only answer using the following sentences: \\
    & A) The persuadee has explicitly refused. \\
    & B) The persuadee remains neutral about donating. \\
    & C) The persuadee has a positive attitude towards donating but hasn't decided yet. \\
    & D) The persuadee has decided to donate. \\
    & Answer:  \\
    \bottomrule
    \end{tabular}
    }
    \caption{Prompts for critic model on P4G.}
    \label{tab:p4g_critic_prompt}
\end{table*}

\begin{table*}[!t]
    \centering\small
    \resizebox{\linewidth}{!}{
    \begin{tabular}{ll}
    \toprule
    System & Now enter the role-playing mode. In the following conversation, you will play as a therapist in a coun-\\
    & seling conversation with a patient. \\
    \midrule
    User & You are the therapist who is trying to help the patient reduce their emotional distress and help them \\
    & understand and work through the challenges. Please reply with only one short and succinct sentence. \\
    & Are you ready to play the game? \\
    Assistant & Yes, I’m ready to play the game! \\
    \midrule
    User & [situation] \\
    \bottomrule
    \end{tabular}
    }
    \caption{Standard Prompts for assistant simulation on ExTES.}
    \label{tab:extes_sys_simu_prompt_standard}
\end{table*}

\begin{table*}[!t]
    \centering\small
    \resizebox{\linewidth}{!}{
    \begin{tabular}{ll}
    \toprule
    System & Now enter the role-playing mode. In the following conversation, you will play as a Persuader who is \\
    & trying to persuade the Persuadee to donate to a charity called Save the Children. Save the Children is \\
    & headquartered in London, and they work to help fight poverty around the world. Children need help in \\
    & developing countries and war zones. Small donations like \$1 or \$2 go a long way to help. \\
    \midrule
    & [conversation] \\
    \midrule
    User & [user response] \\
    & Please reply with only one short and succinct sentence. \\
    \bottomrule
    \end{tabular}
    }
    \caption{Standard Prompts for assistant simulation on P4G.}
    \label{tab:p4g_sys_simu_prompt_standard}
\end{table*}

\begin{table*}[!t]
    \centering\small
    \resizebox{\linewidth}{!}{
    \begin{tabular}{ll}
    \toprule
    System & Assume you are the therapist. Given the conversation history, in order to help the patient reduce their \\
    & emotional distress and help them understand and work through the challenges, please select the most \\
    & appropriate dialogue strategy. \\
    \midrule
    User & You can only reply by selecting one of the following dialogue strategy to reach the goal:\\
    & A) Question \\
    & B) Self-disclosure \\
    & C) Affirmation and Reassurance \\
    & D) Providing Suggestions \\
    & E) Reflection of feelings \\
    & F) Information \\
    & G) Restatement or Paraphrasing \\
    \\
    & The following is the conversation history: [conversation] \\
    & Question: Which one is the most appropriate dialogue strategy? \\
    & Answer: \\
    \bottomrule
    \toprule
    System & Now enter the role-playing mode. In the following conversation, you will play as a therapist in a coun-\\
    & seling conversation with a patient. \\
    \midrule
    User & You are the therapist who is trying to help the patient reduce their emotional distress and help them \\
    & understand and work through the challenges. Please reply with only one short and succinct sentence. \\
    & [action prompt] Are you ready to play the game? \\
    Assistant & Yes, I'm ready to play the game! \\
    \bottomrule
    \end{tabular}
    }
    \caption{Proactive prompts for assistant simulation on ExTES.}
    \label{tab:extes_sys_simu_prompt_proactive}
\end{table*}

\begin{table*}[!t]
    \centering\small
    \resizebox{\linewidth}{!}{
    \begin{tabular}{ll}
    \toprule
    System & Assume you are the Persuader. Given the conversation history, in order to convince the persuadee to donate \\
    & for charity, please select the most appropriate dialogue strategy. \\
    \midrule
    User & You can only reply by selecting one of the following dialogue strategies to reach the goal:\\
    & A) Logical appeal \\
    & B) Emotion appeal \\
    & C) Credibility appeal \\
    & D) Task-related inquiry \\
    & E) Proposition of donation \\
    & F) Greeting.\\
    \\
    & The following is the conversation history: [conversation] \\
    & Question: Which one is the most appropriate dialogue strategy? \\
    & Answer: \\
    \bottomrule
    \toprule
    System & Now enter the role-playing mode. In the following conversation, you will play as a Persuader who is trying \\
    & to persuade the Persuadee to donate to a charity called Save the Children. Save the Children is headquartered \\
    & in London, and they work to help fight poverty around the world. Children need help in developing countries \\
    & and war zones. Small donations like \$1 or \$2 go a long way to help. \\
    \midrule
    & [conversation] \\
    \midrule
    User & [user response]\\
    & Please reply with only one short and succinct sentence. [action prompt] \\
    \bottomrule
    \end{tabular}
    }
    \caption{Proactive prompts for assistant simulation on P4G.}
    \label{tab:p4g_sys_simu_prompt_proactive}
\end{table*}

\begin{table*}[!t]
    \centering\small
    \resizebox{\linewidth}{!}{
    \begin{tabular}{ll}
    \toprule
    System & Assume you are the therapist. Given the conversation history, in order to help the patient reduce their \\
    & emotional distress and help them understand and work through the challenges, please analyze the current \\
    & therapy progress and the patient's emotional state in a concise summary. \\
    & The following is the conversation history: [conversation] \\
    & Please generate a short and succinct analysis. \\
    \bottomrule
    \toprule
    System & Assume you are the therapist. Given the conversation history and concise analysis on this conversation, in \\
    & order to help the patient reduce their emotional distress and help them understand and work through the \\
    & challenges, please select only one of the following dialogue strategies: Question, Self-disclosure, Affirma- \\
    & tion and Reassurance, Providing Suggestions, Reflection of feelings, Information, Restatement or Paraphra-\\
    & sing. \\
    \midrule
    User & The following is the conversation history: [conversation] \\
    & And the following is the concise analysis on this conversation: [analysis] \\
    & Question: Which one is the most appropriate dialogue strategy? \\
    & You can only answer using the following strategies:\\
    & A) Question \\
    & B) Self-disclosure \\
    & C) Affirmation and Reassurance \\
    & D) Providing Suggestions \\
    & E) Reflection of feelings \\
    & F) Information \\
    & G) Restatement or Paraphrasing.\\
    & Answer: \\
    \bottomrule
    \toprule
    System & Now enter the role-playing mode. In the following conversation, you will play as a therapist in a counseling \\
    & conversation with a patient. \\
    \midrule
    User & You are the therapist who is trying to help the patient reduce their emotional distress and help them under-\\
    & stand and work through the challenges. Please reply with only one short and succinct sentence. [action \\
    & prompt] Are you ready to play the game? \\
    Assistant & Yes, I'm ready to play the game! \\
    \bottomrule
    \end{tabular}
    }
    \caption{ProCoT prompts for assistant simulation on ExTES.}
    \label{tab:extes_sys_simu_prompt_procot}
\end{table*}

\begin{table*}[!t]
    \centering\small
    \resizebox{\linewidth}{!}{
    \begin{tabular}{ll}
    \toprule
    System & Assume you are the Persuader. Given the conversation history, in order to convince the persuadee to donate \\
    & for charity, please analyze the current persuasion progress and the persuadee's emotional state in a concise \\
    & summary. \\
    & The following is the conversation history: [conversation] \\
    & Please generate a short and succinct analysis. \\
    \bottomrule
    \toprule
    System & Assume you are the Persuader. Given the conversation history and concise analysis on this conversation, in \\
    & order to convince the persuadee to donate for charity, please select only one of the following dialogue stra-\\
    & tegies: Logical appeal, Emotion appeal, Credibility appeal, Task-related inquiry, Proposition of donation, \\
    & Greeting. \\
    \midrule
    User & The following is the conversation history: [conversation]\\
    & And the following is the concise analysis on this conversation: [analysis]\\
    & Question: Which one is the most appropriate dialogue strategy?\\
    & You can only answer using the following strategies:\\
    & A) Logical appeal\\
    & B) Emotion appeal\\
    & C) Credibility appeal\\
    & D) Task-related inquiry\\
    & E) Proposition of donation\\
    & F) Greeting \\
    & Answer: \\
    \bottomrule
    \toprule
    System & Now enter the role-playing mode. In the following conversation, you will play as a Persuader who is trying \\
    & to persuade the Persuadee to donate to a charity called Save the Children. Save the Children is headquartered \\
    & in London, and they work to help fight poverty around the world. Children need help in developing countries \\
    & and war zones. Small donations like $1 or $2 go a long way to help. \\
    \midrule
    & [conversation] \\
    \midrule
    User & [user response] \\
    & Please reply with only one short and succinct sentence. [action prompt] \\
    \bottomrule
    \end{tabular}
    }
    \caption{ProCoT prompts for assistant simulation on P4G.}
    \label{tab:p4g_sys_simu_prompt_procot}
\end{table*}

\begin{table*}[!t]
    \centering\small
    \resizebox{\linewidth}{!}{
    \begin{tabular}{ll}
    \toprule
    System & Now enter the role-playing mode. In the following conversation, you will play as a coach in a counseling \\
    & game. There will be a therapist and a patient talking about some emotional issues. Your task is to read the \\
    & conversation between the therapist and the patient, then provide suggestions to the therapist about how to \\
    & help the patient reduce their emotional distress and help them understand and work through the challenges. \\
    \midrule
    User & Read the following conversation between the therapist and the patient, then give three suggestions to the \\
    & therapist about how to help the patient reduce their emotional distress and help them understand and work \\
    & through the challenges. Each suggestion should be only one short and succinct sentence.\\
    & The following is the conversation: [conversation]\\
    & Question: What are your suggestions?\\
    & Answer: \\
    \bottomrule
    \toprule
    System & Now enter the role-playing mode. In the following conversation, you will play as a therapist in a counseling \\
    & conversation with a patient. \\
    User & You are the therapist who is trying to help the patient reduce their emotional distress and help them under-\\
    & stand and work through the challenges. Please reply with only one short and succinct sentence based on the \\
    & following suggestion from a coach: [suggestions]\\
    & You need to understand the coach's suggestions and generate appropriate responses, but you cannot mention \\
    & the existence of the coach during the conversation. Are you ready to play the game? \\
    Assistant & Yes, I'm ready to play the game! \\
    \bottomrule
    \end{tabular}
    }
    \caption{ICL-AIF prompts for assistant simulation on ExTES.}
    \label{tab:extes_sys_simu_prompt_icl-aif}
\end{table*}

\begin{table*}[!t]
    \centering\small
    \resizebox{\linewidth}{!}{
    \begin{tabular}{ll}
    \toprule
    System & Now enter the role-playing mode. In the following conversation, you will play as a coach in a persuasion \\
    & game. There will be a persuader who is trying to persuade a persuadee for charity donation. There will be \\
    & a persuader who is trying to persuade a persuadee for charity donation. Your task is to read the conver-\\
    & sation between the persuader and the persuadee, then provide suggestions to the persuader about how to \\
    & convince the persuadee to make a donation. \\
    \midrule
    User & Read the following conversation between the persuader and the persuadee, then give three suggestions to \\
    & the persuader about how to convince the persuadee to make a donation. Each suggestion should be only \\
    & one short and succinct sentence. \\
    & The following is the conversation: [conversation] \\
    & Question: What are your suggestions? \\
    & Answer: \\
    \bottomrule
    \toprule
    System & "Now enter the role-playing mode. In the following conversation, you will play as a Persuader who is \\
    & trying to persuade the Persuadee to donate to a charity called Save the Children. Save the Children is head-\\
    & quartered in London, and they work to help fight poverty around the world. Children need help in develo-\\
    & ping countries and war zones. Small donations like \$1 or \$2 go a long way to help. \\
    \midrule
    & [conversation] \\
    \midrule
    User & [user response] \\
    & Please reply with only one short and succinct sentence based on the following suggestion from a coach: \\
    & [suggestions]\\
    & You need to understand the coach's suggestions and generate appropriate responses, but you cannot men-\\
    & tion the existence of the coach during the conversation. \\
    \bottomrule
    \end{tabular}
    }
    \caption{ICL-AIF prompts for assistant simulation on P4G.}
    \label{tab:p4g_sys_simu_prompt_icl-aif}
\end{table*}

\end{document}